%% file: ARXIV-ASTRA-bench.tex
\documentclass{article}

\usepackage{microtype}
\usepackage{graphicx}
\usepackage{subcaption}
\usepackage{booktabs} 
\usepackage{xcolor}         
\usepackage{caption}
\usepackage{listings}
\usepackage{tabularx}
\usepackage{pifont}
\usepackage{adjustbox}
\usepackage{multirow}
\usepackage{makecell}
\usepackage{authblk}
\usepackage{amsmath} 
\usepackage{bbm}
\usepackage[most]{tcolorbox} 
\usepackage{CJKutf8}
\usepackage{hhline}
\usepackage[inline]{enumitem}
\newcommand{\cmark}{\textcolor{green!70!black}{\ding{51}}} 
\newcommand{\xmark}{\textcolor{red}{\ding{55}}}             
\newcommand{\beginsupplement}{%
        \setcounter{table}{0}
        \renewcommand{\thetable}{S\arabic{table}}%
        \setcounter{figure}{0}
        \renewcommand{\thefigure}{S\arabic{figure}}%
     }
\usepackage{hyperref}

\usepackage[preprint]{icml2026}

\usepackage{amsmath}
\usepackage{amssymb}
\usepackage{mathtools}
\usepackage{amsthm}

\usepackage[capitalize,noabbrev]{cleveref}

\theoremstyle{plain}

\theoremstyle{definition}

\theoremstyle{remark}

\usepackage[textsize=tiny]{todonotes}

\icmltitlerunning{ASTRA-bench: Evaluating Tool-Use Agent Reasoning and Action Planning with Personal User Context}

\begin{document}

\twocolumn[
  \icmltitle{ASTRA-bench: Evaluating Tool-Use Agent Reasoning \\ and Action Planning with Personal User Context}



  \icmlsetsymbol{equal}{*}

  \begin{icmlauthorlist}
    \icmlauthor{Zidi Xiu}{equal,aapl}
    \icmlauthor{David Q.~Sun}{equal,aapl,done}
    \icmlauthor{Kevin Cheng}{aapl,done}
    \icmlauthor{Maitrik Patel}{aapl}
    \icmlauthor{Josh Date}{aapl}
    \icmlauthor{Yizhe Zhang}{aapl}
    \icmlauthor{Jiarui Lu}{aapl}
    \icmlauthor{Omar Attia}{aapl}
    \icmlauthor{Raviteja Vemulapalli}{aapl}
    \icmlauthor{Oncel Tuzel}{aapl}
    \icmlauthor{Meng Cao}{aapl}
    \icmlauthor{Samy Bengio}{aapl}
  \end{icmlauthorlist}

  \icmlaffiliation{aapl}{Apple}
  \icmlaffiliation{done}{Work done while at Apple}

\icmlcorrespondingauthor{Zidi Xiu}{z\_xiu@apple.com}

  \icmlkeywords{Data Science and Annotation, Human-Computer Interaction
Benchmark, personal assistant, language agents, planning, Reasoning, Synthetic Data}

  \vskip 0.3in
]



\printAffiliationsAndNotice{}  

\begin{abstract}
Next-generation AI must manage vast personal data, diverse tools, and multi-step reasoning, yet most benchmarks remain context-free and single-turn. We present ASTRA-bench (Assistant Skills in Tool-use, Reasoning \& Action-planning), a benchmark that uniquely unifies time-evolving personal context with an interactive toolbox and complex user intents. Our event-driven pipeline generates 2,413 scenarios across four protagonists, grounded in longitudinal life events and annotated by referential, functional, and informational complexity. Evaluation of state-of-the-art models (e.g., Claude-4.5-Opus, DeepSeek-V3.2) reveals significant performance degradation under high-complexity conditions, with argument generation emerging as the primary bottleneck. These findings expose critical limitations in current agents' ability to ground reasoning within messy personal context and orchestrate reliable multi-step plans. We release ASTRA-bench with a full execution environment and evaluation scripts to provide a diagnostic testbed for developing truly context-aware AI assistants.
\end{abstract}

\section{Introduction}
\vspace{-.5em}
Large language models (LLMs) have demonstrated remarkable general-purpose capabilities across language understanding, reasoning, and generation tasks \citep{brown2020language}.  Within the \emph{intelligent assistant} (IA) space, early uses of LLMs mainly replaced individual natural language understanding (NLU) components --- such as intent classifiers, parsers, and entity resolvers \citep{chen2019bert}.  More recent systems are taking greater advantage of the LLM’s ability to \emph{understand multi-modal context}\citep{achiam2023gpt}, \emph{plan}, \emph{reason}, and \emph{manage conversational state} \citep{yao2022react} over long horizons \citep{park2023generative}.  As users increasingly rely on digital assistants in everyday workflows, building agents that can autonomously achieve complex user goals becomes both technically urgent and economically valuable.

\begin{figure*}[ht]
    \centering
    \includegraphics[width=.95\linewidth]{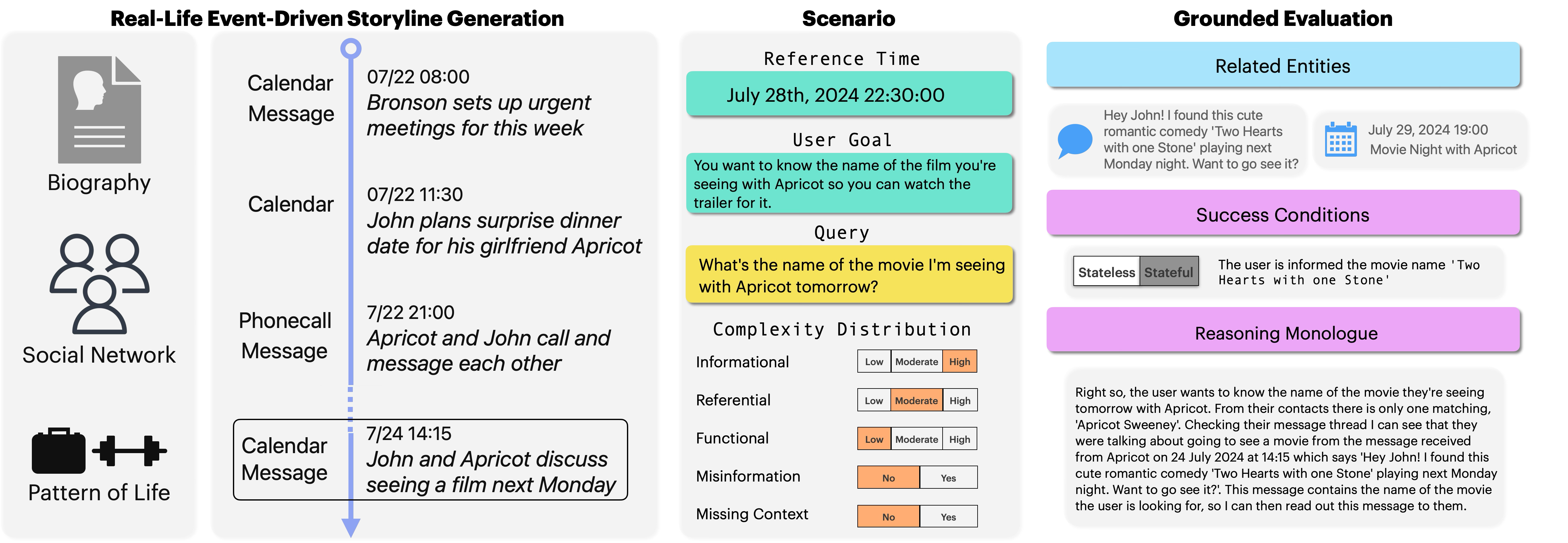}
    \caption{\small{The ASTRA-bench Evaluation Lifecycle. Scenarios are grounded in a protagonist’s behavioral prior and time-evolving storyline. Events are projected into multi-source personal data (Panel 1 \& 2), forming the basis for human-authored queries annotated by complexity tiers. Evaluation is grounded in intermediate reasoning monologues, related entities, and verifiable success conditions (Panel 3 \& 4).}}
    \label{fig:astra-overview}
    \vspace{-1.5em}
\end{figure*}

\noindent\textbf{Desiderata for next-generation assistants.} We distill three core capabilities that any truly capable personal assistant ought to master: 
\begin{enumerate*}[label=(\arabic*)]
    \item \textbf{Context inference \& understanding}: ground user queries in fine-grained personal, system, and world context;
    \item \textbf{Robust reasoning with tool use}: decide \emph{when}, \emph{how}, and \emph{in what order} to invoke external tools while gracefully handling failures;
    \item \textbf{Iterative action planning}: decompose high-level goals into executable steps, maintaining state across turns, and adapting plans as new information arrives \citep{yao2022react, schick2023toolformer, park2023generative}.
\end{enumerate*}
Fulfilling these desiderata in practice requires operating at the intersection of three information sources: 
\begin{enumerate*}[label=(\roman*)]
\item  the device’s system capabilities (installed apps, available APIs, sensors); 
\item rich, longitudinal personal-context data (emails, calendars, notes, files, preferences); 
and 
\item the dynamically changing state of the external world \citep{thoppilan2022lamda, park2023generative}. 
\end{enumerate*}
Crucially, it is the interaction among these sources—rather than any single factor in isolation—that gives rise to the most challenging real-world assistant behaviors.
Yet most existing benchmarks factorize these challenges, evaluating tool use, context grounding, or planning in isolation, and therefore fail to capture the compounded difficulty of realistic, stateful assistant use.

A growing body of benchmarks captures individual aspects of these challenges.
For example, GTA \citep{wang2024gta} evaluates general tool agents using human-written real user queries together with real deployed tools and authentic multimodal contexts;
UltraTool \citep{huang2024planning} emphasizes the end-to-end process of tool utilization, including explicit evaluation of planning, tool creation, and tool usage in complex scenarios;
DialogTool \citep{wang2025rethinking} targets stateful tool use in multi-turn dialogues across the full tool lifecycle and introduces a virtual mobile environment for API execution;
and Vending-Bench \citep{backlund2025vending} stress-tests long-horizon coherence in a persistent simulated environment.
Meanwhile, large-scale API-centric resources such as ToolLLM focus on tool invocation over thousands of real-world APIs, largely via automatically constructed instructions and solution paths \citep{qin2023toolllm}.
However, these benchmarks typically prioritize specific facets (e.g., general tool execution, tool creation, dialogue-stateful tool use, or long-horizon coherence) rather than jointly modeling \emph{longitudinal personal-context corpora} (e.g., emails/calendars/messages), \emph{realistic personal-assistant tool suites}, and \emph{diagnostic complexity axes} within a unified, stateful evaluation protocol.

Furthermore, benchmarks such as GAIA-2 \citep{andrews2025scaling} evaluate tool-augmented agents through multi-step tasks of varying difficulty, but largely treat task complexity as a scalar property. In contrast, ASTRA-bench decomposes task difficulty into orthogonal dimensions capturing referential ambiguity, informational demands, and functional planning requirements, enabling more diagnostic evaluation of agent capabilities.

Consequently, such benchmarks often fail to reveal the unique challenges that arise when an agent must reason over a user’s noisy, complex, multi-source context while orchestrating real tools.  
A next-generation benchmark should therefore jointly provide:
\begin{itemize}[nosep, leftmargin=1.5em]
  \item \emph{Coherent, diverse personal-context traces} spanning weeks or months;
  \item \emph{Executable tool environments} that persist across dialogue turns;
  \item \emph{Diverse user queries} with reasoning monologues and success conditions.
\end{itemize}

To bridge this gap, we introduce \textbf{ASTRA-bench} - \underline{A}ssistant \underline{S}kills in \underline{T}ool-use, \underline{R}easoning, and \underline{A}ction-planning.  ASTRA-bench couples
\begin{enumerate*}[label=(\roman*)]
    \item time-evolving personal-context data grounded in realistic life events,
    \item an interactive tool sandbox (email, calendar, messages, \emph{etc.}), 
and \item 2,413 human-authored conversational scenarios with \emph{context-aware} goals at varying complexity levels along referential, functional, and informational axes. 
\end{enumerate*}
An overview of the benchmark is shown in Fig.~\ref{fig:astra-overview}.
ASTRA-bench builds upon ToolSandbox \citep{lu2024toolsandbox}, extending it with longitudinal time awareness, richer personal context, and structured complexity annotations.
In Table~\ref{tab:benchmarks}, we compare ASTRA-bench against existing frameworks based on five key axes. Specifically, Human Authorship (Hum. Auth.) distinguishes scenarios with explicit human intent from purely synthetic templates, while Trajectory Verifiability (Traj. Verif.) ensures that models are evaluated on intermediate tool execution traces rather than just final outputs.

To support \emph{scalable yet reliable} evaluation, ASTRA-bench adopts a grounded evaluation framework that leverages observable evidence—such as tool traces and system state—to enable fine-grained, diagnostic assessment of agent behavior beyond final-answer correctness.

Using ASTRA-bench, we perform a zero-shot study of state-of-the-art proprietary and open-source LLMs equipped with tool-calling.  Our results show that even state-of-the-art reasoning-tuned models are still insufficient to act as an effective personal assistant. 



\vspace{-1.2em}
\paragraph{Summary of Contributions}
\vspace{-3pt}
\begin{description}[nosep, leftmargin=0pt, font=\bfseries]
    \item[1. ASTRA-bench:] A benchmark with 2,413 scenarios grounded in longitudinal life events with multi-dimensional complexity labels. 
    \item[2. Evaluation Protocol:] Grounded assessment framework leveraging observable evidence for diagnostic analysis of agent behavior.
    \item[3. Empirical Study:] Revealing bottlenecks in argument generation and performance decay under high complexity.
\end{description}

\nocite{mok2025exploring, trivedi2024appworld, yao2024tau}

\vspace{-0.8em}
\section{Related Work}
\vspace{-3pt}
\subsection{Tool-Use Agents and Benchmarks}
\vspace{-2pt}
Early tool-use agents focused on augmenting LLMs with external knowledge via web browsing \citep{nakano2021webgpt} or self-supervised API discovery \citep{schick2023toolformer}. As core challenges shifted from simple tool selection to integrating tool outputs into coherent, multi-step responses, frameworks like ReAct~\citep{yao2022react}, ToolLLM~\citep{qin2023toolllm}, and Gorilla~\citep{patil2025bfcl} improved reasoning loops and invocation accuracy. This evolution has moved toward orchestrator-style agents and user-centric assistants using realistic context for personalized tasks.

Parallel to agent development, benchmarks have evolved from single-turn function-calling (e.g., API-Bank~\citep{li2023api}, BFCL~\citep{patil2025bfcl}) to multi-step reasoning tasks like AgentBench~\citep{liu2023agentbench} and GAIA~\citep{mialon2023gaia}. While these broaden task coverage, they often assess interactions in isolation using only outcome-based criteria. Recent efforts like GAIA-2~\citep{andrews2025scaling}, ToolTalk~\citep{farn2023tooltalk}, and $\tau^2$-Bench~\citep{barres2025tau} introduce multi-turn goals but largely rely on outcome-based success criteria.

{\setlength{\textfloatsep}{6pt}
 \setlength{\intextsep}{6pt}
\begin{table}[htbp!]
\centering
\begin{small}
\begin{sc}
\begin{adjustbox}{max width=.9\columnwidth}
\begin{tabularx}{\linewidth}{l*{5}{>{\centering\arraybackslash}X}}
\toprule
\textbf{Benchmark} & \textbf{\shortstack{Hum.\\Auth.}} & \textbf{\shortstack{Traj.\\Verif.}} & \textbf{\shortstack{Skill\\Comp.}} & \textbf{\shortstack{Ctx.\\Goals}} & \textbf{\shortstack{Gnd.\\Ctx.}} \\
\midrule
\textbf{ASTRA-bench}        & \cmark & \cmark & \cmark & \cmark & \cmark \\
HiCUPID               & \xmark & \xmark & \xmark & \cmark & \xmark \\
AppWorld              & \cmark & \cmark & \cmark & \xmark & \cmark \\
ToolSandbox           & \cmark & \cmark & \cmark & \xmark & \xmark \\
$\tau^2$-Bench          & \cmark & \cmark & \xmark & \xmark & \xmark \\
ToolTalk              & \cmark & \cmark & \cmark & \xmark & \xmark \\
AgentBench            & \xmark & \cmark & \xmark & \xmark & \xmark \\
GAIA                  & \cmark & \xmark & \cmark & \xmark & \xmark \\
GAIA-2                & \cmark & \xmark & \cmark & \cmark & \cmark \\
API-Bank              & \cmark & \cmark & \cmark & \xmark & \xmark \\
BFCL                  & \cmark & \cmark & \cmark & \xmark & \xmark \\
\bottomrule
\end{tabularx}
\end{adjustbox}
\end{sc}
\end{small}
\caption{
Comparison of tool-use benchmarks. Criteria include Human Authorship (Hum. Auth.), Trajectory Verifiability (Traj. Verif.), Skill Complexity (Skill Comp.), Context-Dependent Goals (Ctx. Goals), and Grounded Personal Context (Gnd. Ctx.).
}
\label{tab:benchmarks}
\vspace{-10pt}
\end{table}}
\vspace{-7pt}
\subsection{Personal Context in Agent Evaluation}
\vspace{-3pt}
A critical frontier is evaluating agents within persistent, stateful environments. AppWorld~\citep{trivedi2024appworld} provides a mobile-app sandbox but focuses on code generation rather than structured function calls. ToolSandbox~\citep{lu2024toolsandbox} and HiCUPID~\citep{mok2025exploring} move toward stateful settings where user state evolves, with the latter targeting personalization capabilities.

As shown in Table~\ref{tab:benchmarks}, ASTRA-bench fills this gap by providing an interactive tool sandbox, covering email, calendars, and messaging, coupled with a grounded evaluation framework. Unlike previous benchmarks that rely on surface-level text correctness, we leverage observable evidence from tool traces and system state snapshots to enable fine-grained, diagnostic assessment of agent behavior. This ensures the agent is evaluated on its ability to navigate complex, multi-source personal context rather than merely matching an end-goal answer.

\vspace{-12pt}
\subsection{Synthetic Personas and Storylines}
\vspace{-5pt}

LLMs are increasingly used to generate synthetic personal data to reduce reliance on costly human annotation~\citep{chen2023places}. Early work like the Persona-Chat corpus grounded dialogues in predefined personas~\citep{zhang2018personalizing}, which has since been scaled using generator-critic frameworks to preserve fidelity. Recent efforts, such as Conversation Chronicles~\citep{jang2023conversation} and HiCUPID~\citep{mok2025exploring}, have expanded this to multi-session dynamics and rich user metadata—like schedules and traits—that unfold across dialogue turns. Despite these advances, existing datasets often treat interactions as self-contained episodes, which limits their ability to evaluate an agent's long-term continuity. ASTRA-bench extends this paradigm by shifting from episodic personas to ``Protagonists'' whose digital artifacts are grounded in a continuous, time-evolving storyline. As detailed in Section~\ref{sec:personal-context}, we utilize an event-driven generation pipeline to ensure that every email, calendar entry, and message jointly projects a coherent, longitudinal narrative, providing a more realistic testbed for persistent AI assistants.

\vspace{-1em}
\section{Personal Data Generation}\label{sec:personal-context}
\vspace{-3pt}

Personal-context data grounded in realistic life events—rather than unconstrained raw human logs—form the core of ASTRA-bench.
An assistant expected to reschedule ``Dinner with Jess'' or find a ``book recommendation from my brother'' cannot succeed unless the underlying text messages, calendar entries, contacts, and emails jointly tell a believable story. We therefore propose a pipeline that synthesizes personal user data by first generating a sequences of events and then producing the associated personal data.
\vspace{-8pt}
\subsection{The Protagonist: Real-Life Event-Driven Storyline}
\vspace{-5pt}

\begin{figure*}[ht]
\vspace{-5pt}
    \centering
    \includegraphics[width=0.8\linewidth]{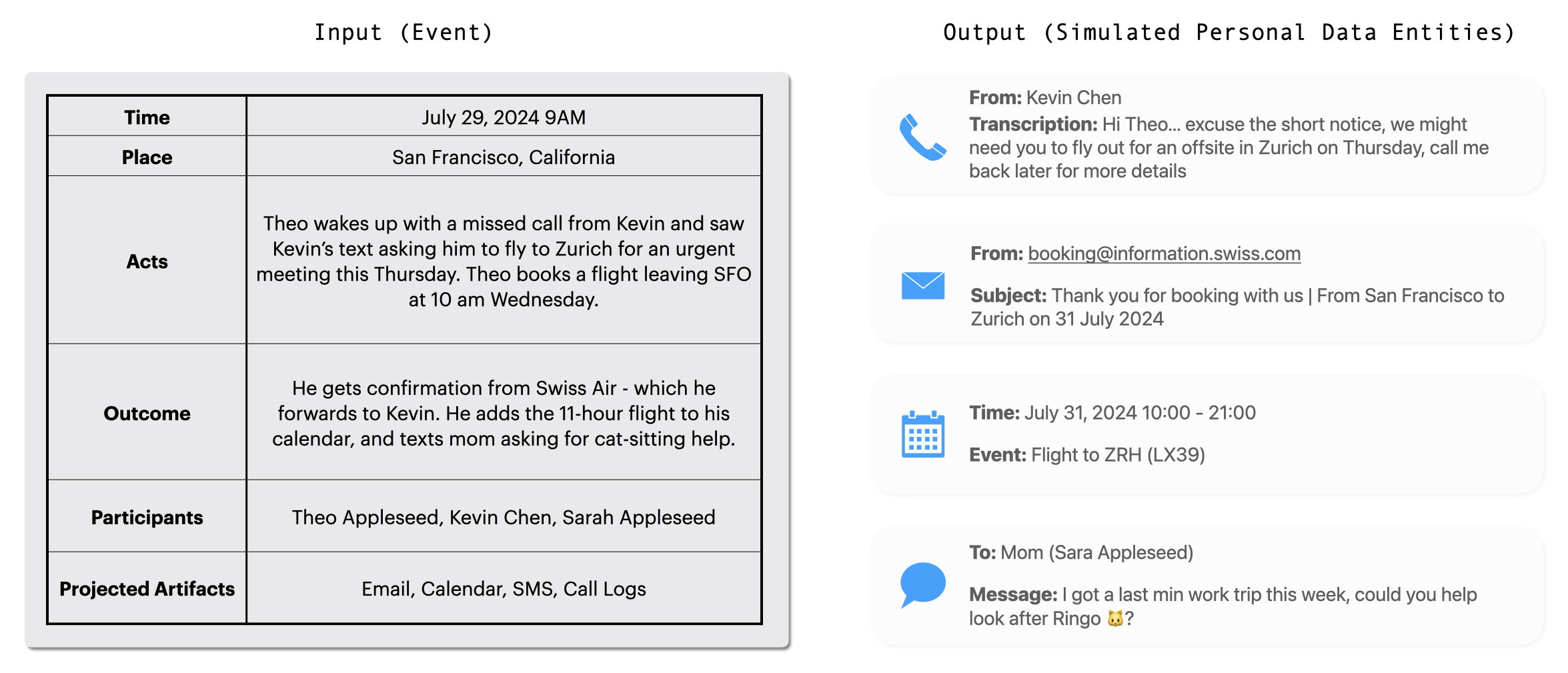}
    \vspace{-5pt}
    \caption{Example of an Event and its Projected Artifacts from our Generation Workflow}
    \label{fig:so-io-example}
    \vspace{-1.5em}
\end{figure*}

To ensure longitudinal consistency, inspired by the Agent-Based Modeling paradigm \citep{silverman2006abm} and drama-writing principles \citep{spencer2002playwright}, we introduce the concept of a Protagonist—the simulated owner of the digital environment. 
Unlike previous benchmarks that generate tasks in isolation, our pipeline grounds all digital artifacts in a coherent ``behavioral prior'' composed of three layers:
\begin{description}[nosep, leftmargin=0pt, font=\bfseries]
  \item[Biography \& Social Network:] Each protagonist is instantiated with a fixed biography (e.g., demographic fields, socioeconomic markers, long-term preferences) and a directed multigraph representing their social network. This graph encodes relationship types (Family, Friend) and contact frequencies, ensuring that simulated interactions are grounded in stable social dynamics. 
  \item[Pattern of Life (PoL):] We implement a probabilistic context-free grammar to generate daily agendas and weekly routines. The PoL expands high-level routines—such as commutes or family time—into timestamped events, capturing realistic circadian rhythms and weekday/weekend asymmetries.
  \item[Long-Horizon Consistency:] By conditioning downstream events on this frozen biography and social graph, we produce a ``storyline'', a sequence of persistent events anchored in character habits and obligations. This ensures digital interactions remain consistent over weeks or months.
\end{description}

\vspace{-8pt}
\subsection{The Tenet: Event-Driven Context Data Generation} \label{sec:tenet}
\vspace{-5pt}
A central premise to our design is that every digital artifact is a lossy projection of an \emph{event} in the real world. By first simulating events, we can then ``project'' those events onto application-specific records. 

We represent each event as a structured object that anchors the simulation. This schema provides a clear, structured input for our generation pipeline and is defined by:
\begin{description}[nosep, leftmargin=0pt, font=\bfseries]
  \item[Contextual Anchors:] Absolute Time (anchoring the notion of ``now") and Place (physical or virtual location).
  \item[Narrative Core:] Acts (primary actions) and Outcome (triggered follow-up actions).
  \item[Social \& Digital Trace:] Participants (social network members involved) and Projected Artifacts (the resulting digital entities, such as emails, calendar invites, or SMS logs).
\end{description}
By repeating this projection for every event across the storyline, we ensure that a single occurrence—such as a theatre outing—consistently manifests as a shared calendar invite, an email confirmation, and related text messages.
A sample input (event) and output (expected personal data entities) is shown in Figure~\ref{fig:so-io-example}.

\vspace{-10pt}
\subsection{Agentic Generation Workflow}
\vspace{-5pt}
We translate each high-level \emph{event} into concrete multi-app artifacts via a cascade of lightweight, task-specialized \emph{agents}.  
The pipeline follows a structured  
\textbf{draft $\rightarrow$ critique $\rightarrow$ revise $\rightarrow$ verify} loop that fuses the generative fluency of LLMs with systematic self-reflection:
\begin{description}[nosep, leftmargin=1em, font=\small\bfseries]

  \item \textbf{Draft Agent (Generator)}  
        receives the structured event object and emits a first-pass set of candidate artifacts  
        (e.g. e-mail body, calendar metadata, SMS text).
  \item \textbf{Critique Agent (Self-Reviewer)}  
        checks the draft against domain-specific acceptance criteria (e.g. factual coherence with the event, stylistic consistency with the protagonist, schema validity, and cross-app coherence) and returns a structured list of violations.
  \item \textbf{Revision Agent (Fixer)}  
        consumes the critique and produces an improved artifact set.  
        Localized failures allow revisions to typically converge within one or two iterations.
  \item \textbf{Verification Agent (Gatekeeper)}  
        runs schema validation, referential-integrity checks, and temporal consistency tests.  
        Outputs are either \textsc{accept} or a minimal failure report that re-enters the critique–revision loop.
\end{description}

Each agent is defined only by 
\begin{enumerate*}[label=(\roman*)]
    \item a concise task description,
    \item its reasoning rubric, and
    \item  machine-checkable outputs—no additional fine-tuning is required.  
\end{enumerate*}
Agents can be swapped or extended (e.g., inserting a \textit{Redaction Agent} to remove personally identifiable information) without retraining the pipeline.  
Because only failing components re-run, the cascade amortises compute: fewer than 8\,\% of events trigger a second revision pass in our experiments.

\vspace{-8pt}
\section{Protagonist-Centered Scenarios Authoring}
\vspace{-3pt}

Building on the artifacts from Section~\ref{sec:personal-context}, we now define the user-agent interaction framework. We translate storylines into executable \emph{scenarios}  (interchangeable with \emph{tasks}), where goals are grounded in specific life events. As shown in Figure~\ref{fig:astra-overview}, every query—from rescheduling dinner to finding flights—becomes a natural extension of the protagonist's digital life rather than a context-free instruction. This grounding provides the verifiable foundation for diagnostic evaluation of reasoning and multi-turn planning.

\vspace{-5pt}
\subsection{User-Centric Scenario Set Up}
\vspace{-3pt}
\textbf{User goal}. 
A user goal specifies the objective the protagonist intends to achieve. Rather than using static prompts, these goals guide an LLM-powered user simulator in multi-turn dialogues, where the agent may need to seek clarification if the initial request is ambiguous. To maintain realism, each simulator is constrained by a knowledge boundary, ensuring it only shares information the protagonist would naturally know, thereby forcing the agent to rely on tool interaction for fact-finding.

\textbf{Reference time}. Crucially, every goal is anchored by a reference time that defines ``now'' within the protagonist’s longitudinal data. Because personal context (e.g., ``unread'' emails or ``upcoming'' meetings) is time-dependent, the agent must perform temporal canonicalization. It must transform relative expressions like ``next Friday'' or ``last night'' into precise timestamps to filter records accurately. This temporal awareness is a primary differentiator from context-free benchmarks, as it requires agents to distinguish between historical routines and current obligations.

\begin{table*}[ht!]
\vspace{-5pt}
\centering
\begin{small}
\begin{sc}
\caption{ASTRA-bench dataset composition (top) and complexity distribution (bottom)}
\label{tab:ASTRA-bench-dist}
\resizebox{.8\textwidth}{!}{
\begin{tabular}{l|ccc|cccccc}
\toprule
\multirow{2}{*}{\textbf{Count}} &
\multicolumn{3}{c|}{\textbf{Scenarios Count}} &
\multicolumn{6}{c}{\textbf{Domain Context Count}} \\
\cline{2-10}
& \rule{0pt}{2.2ex}\textbf{Events} & \textbf{User Goals} & \textbf{Scenarios}
& \textbf{Contact} & \textbf{Calendar} & \textbf{Email} & \textbf{Message} & \textbf{WhatsApp} & \textbf{Phone call} \\
\hline
\rule{0pt}{2.2ex}\textbf{Total} &
111 & 1360 & 2413 &
69 & 115 & 84 & 309 & 30 & 15 \\
\bottomrule
\end{tabular}
}

\vspace{0.3ex}

\resizebox{.8\textwidth}{!}{
\begin{tabular}{l|ccc|ccc|ccc|cc|cc}
\toprule
\multicolumn{14}{c}{\textbf{Complexity Distribution}} \\
\midrule
\multirow{2}{*}{\textbf{}} &
\multicolumn{3}{c|}{\textbf{Referential Complexity}} &
\multicolumn{3}{c|}{\textbf{Informational Complexity}} &
\multicolumn{3}{c|}{\textbf{Functional Complexity}} &
\multicolumn{2}{c|}{\textbf{Misinformation}} &
\multicolumn{2}{c}{\textbf{Insufficient Context}} \\
\cline{2-14}
\rule{0pt}{2.2ex} & \textbf{Low} & \textbf{Moderate} & \textbf{High}
& \textbf{Low} & \textbf{Moderate} & \textbf{High}
& \textbf{Low} & \textbf{Moderate} & \textbf{High}
& \textbf{No} & \textbf{Yes}
& \textbf{No} & \textbf{Yes} \\
\hline
\rule{0pt}{2.2ex} \textbf{Count} &
390 & 1733 & 290 &
642 & 1292 & 479 &
1386 & 857 & 170 &
2376 & 37 &
2396 & 17 \\
\bottomrule
\end{tabular}
}
\vspace{-12pt}
\end{sc}
\end{small}
\end{table*}

\vspace{-10pt}
\subsection{User Queries and Complexity Tiers}\label{sec:complexity-tiers}
\vspace{-5pt}
To enable fine-grained diagnostic analysis, we annotate each scenario across three orthogonal complexity axes. This multi-dimensional stratification (visualized in Figure~\ref{fig:astra-overview}) allows us to isolate specific failure modes in agent reasoning, retrieval, and tool-calling.

\vspace{-10pt} 
\paragraph{Informational complexity.} 
This axis measures the depth of reasoning and multi-step synthesis required to satisfy the query. \textit{Low}-complexity tasks involve direct retrieval, while \textit{high}-complexity tasks require synthesizing information across multiple apps or longitudinal time spans (e.g., ``Find the restaurant I visited with the same person I met for coffee last Tuesday'').

\vspace{-10pt}
\paragraph{Referential complexity.} 
This axis quantifies the effort required to ground ambiguous or indirect mentions into concrete entities. \textit{High} referential complexity involves resolving multiple underspecified pronouns (``her'', ``them'') or time-anchored descriptions (``my last guest'') that require cross-referencing the protagonist's social graph and communication history.

\vspace{-10pt}
\paragraph{Functional complexity.} 
This axis tracks the coordination and logic of the required tool calls. It evaluates the agent’s ability to handle conditional dependencies, parallel tool execution, and the correct sequencing of actions within the stateful environment.

\vspace{-10pt}
\paragraph{Robustness stress tests.} 
Beyond standard execution, we introduce two critical stress conditions: \emph{Misinformation} and \emph{Insufficient Context}. In misinformation scenarios, the agent must detect and reconcile conflicting or erroneous data within the personal context. In insufficient context scenarios, the agent must recognize that the user's request is underspecified and proactively seek clarification via the user simulator.

\vspace{-10pt}
\subsection{Components of Annotated Ground Truth for Scenario Evaluation}
\vspace{-5pt}

To support grounded evaluation, each scenario is paired with a comprehensive ground-truth annotation. This annotation moves beyond simple ``correct answer'' to provide a full trace of the intended reasoning and system state changes.

\textbf{Related entities.} 
We identify the set of personal context (e.g. email threads, contact cards, etc.) that an agent must interact with to fulfill the goal. These entities serve as a ``retrieval gold standard'', allowing us to verify if the agent correctly navigated the protagonist’s complex social and digital history.

\textbf{Reasoning monologue.} 
Each scenario includes a human-authored chain-of-thought that details the optimal logic required to solve the task. This monologue explains how to resolve ambiguities, which tools to prioritize, and how to handle potential data conflicts. It acts as the reference for evaluating the agent’s internal reasoning trajectory.

\textbf{Success conditions.} 
We define the required terminal state of the environment and the specific information that must be conveyed to the user. These conditions represent the ground-truth outcome for each scenario. Success conditions can be (1) stateless, requiring no modification to persistent data (e.g., providing a precise answer), or (2) stateful, involving changes to the environment or device state (e.g., sending a confirmed email).

These components form the foundation of our evaluation frameworks, introduced in Sec.~\ref{sec:evaluation}, which includes both LLM-based judgment and verifiable milestone calculations, following methodologies similar to \citet{lu2024toolsandbox}.

\vspace{-10pt}
\section{ASTRA-bench}
\vspace{-5pt}

\subsection{Scale and Diversity of ASTRA-bench}
\vspace{-3pt}

The ASTRA-bench dataset\footnote{ASTRA-bench Evaluation Framework is released at \url{https://github.com/apple-aiml-research/ml-astra}} consists of 2,400 human-authored scenarios grounded in the digital lives of five unique protagonists. Each protagonist is defined by a distinct demographic and behavioral profile, with digital histories spanning an average of 14 days to ensure long-horizon context.

As summarized in Table~\ref{tab:ASTRA-bench-dist}, the benchmark provides a high-density environment for agent interaction, containing over 600 projected digital artifacts across email, messaging, and calendar applications. By focusing on a deep history for each of the five protagonists, we ensure that agents must navigate a significant volume of realistic ``noise''-including overlapping meetings and recurring social threads—to identify the precise entities required for task completion. The scenarios are balanced across our complexity tiers (Section~\ref{sec:complexity-tiers}) to provide a robust diagnostic signal.

\vspace{-10pt}
\subsection{Simulation Environment}
\vspace{-5pt}

Our simulation environment is built upon the foundational principles of ToolSandbox \citep{lu2024toolsandbox}, but introduces critical enhancements to address the unique challenges of personal assistant evaluation.

\textbf{Personal Context Integration.} While the original framework focuses on generic tool execution, our sandbox is grounded in the protagonist’s digital history. Every tool call interacts with the 622 projected artifacts generated in our workflow, ensuring that the agent must reason over specific personal data—such as social graphs and past communications—to fulfill a request.

\textbf{Temporal Awareness.} A core advancement in our environment is the introduction of a global reference time. This ``temporal anchor'' forces the agent to perform canonicalization of relative time expressions (e.g., \emph{yesterday} or \emph{last week}) against the persistent system state. This ensures that tool outputs are factually synchronized with the timeline of the protagonist’s life.

\textbf{Expanded Tool Coverage} We significantly expand the breadth of available tools to include 25+ more across six domains: Contact, Calendar, Email, Message, WhatsApp, and Phone Call. This cross-app coverage requires the agent to coordinate actions across multiple digital silos, moving beyond the single-app focus of many existing benchmarks.

\textbf{Enhanced Evaluation Strategies.} We provide human-authored evaluation components that measure task success relative to explicit user goals, leveraging success conditions, reasoning monologues, and related entities, moving beyond the Milestones and Minefields introduced in ToolSandbox. Further, milestone and minefield evaluations are refined to emphasize information retrieval accuracy and user-facing outcomes, moving beyond strict tool-parameter matching. In addition, an LLM Judge is incorporated to assess task success using these same ground-truth components, offering a complementary evaluation that captures retrieval and reasoning quality.

\vspace{-8pt}
\subsection{Evaluation Methodology}\label{sec:evaluation}
\vspace{-5pt}
We evaluate performance by comparing agent trajectories against human-authored ground-truth targets (goals, queries, success conditions, and entities). Assessment is grounded in \emph{observable evidence}—tool traces and system state snapshots—enabling multi-dimensional diagnostics through two decoupled methodologies.

\paragraph{Verifiable Measures: Milestones and Minefields.}
We adopt and extend the Milestones \& Minefields framework \citep{lu2024toolsandbox} to capture intermediate, fact-grounded sub-goals such as retrieval accuracy and payload correctness. Unlike the original framework, we structure these as a directed acyclic graph (DAG) to enforce temporal and logical dependencies, denoted as $G_{M^+}(V_{M^+}, E_{M^+})$ for milestones and 
$G_{M^-}(V_{M^-}, E_{M^-})$ for minefields. We further introduce high-resolution similarity metrics to specifically evaluate information retrieval accuracy and goal alignment. A critical feature of our stateful environment is the enforcement of a strict penalty: any minefield violation—such as the unintended removal of calendar events—nullifies the overall score, even if all milestones are achieved.
\begin{equation}\label{eq:rule-based-score}
S_{\text{Rule-Based}} = S_{\text{Milestones}} \times \mathbbm{1}{[S_{\text{Minefields}}=0]}
\end{equation}
These measures are initially generated by GPT-o3 based on grounded evaluation from annotation and subsequently revised by humans to ensure diagnostic precision.

\paragraph{LLM-Based Evaluators.} 
In parallel, rubric-guided LLM judges assess the same evidence alongside human-authored rationales and personal context. This captures higher-level signals like conversational efficiency and hallucination robustness. Performance is scored (0–2) across five dimensions: \emph{Task Completion}, \emph{Tool Usage}, \emph{Information Retrieval}, \emph{Conversation Effectiveness}, and \emph{No Hallucination}. This provides a nuanced assessment of reasoning and semantic correctness where rule-based metrics may be too rigid.

\paragraph{Comparative Perspective.} While rule-based measures are precise but rigid, LLM judges offer flexible, interpretive assessments. Despite these differences, model rankings across both methods show strong alignment. This dual-approach captures both structured success and broader reasoning quality (see Appendix for detailed comparison).

\vspace{-10pt}
\section{Results}
\vspace{-5pt}
\subsection{Experiment Setup}


\paragraph{Models.}  
We evaluate a diverse suite of state-of-the-art models, ranging from frontier proprietary systems to leading open-source checkpoints, as detailed in Table~\ref{tab:models}. This includes frontier systems like Claude-4.5-Opus and GPT-4.1, alongside the DeepSeek-V3.2 and Qwen3 families, providing a comprehensive cross-section of current reasoning architectures and scales.
\begin{table}[h]
\centering
\small
\setlength{\tabcolsep}{4pt} 
\begin{small}
\begin{sc}
\begin{tabularx}{\columnwidth}{l X}
\toprule
\textbf{Category} & \textbf{Model Variants \& Citations} \\ 
\midrule
\textbf{Proprietary} & GPT-o3, GPT-4.1 \citep{hurst2024gpt}; Claude-4.5-Opus, Claude-4.5-Haiku \citep{anthropic_claude} \\
\addlinespace[2pt]
\textbf{Open-source} & DeepSeek-v3.2 \citep{liu2025deepseek}; Qwen3 (30B-A3B, 30B-A3B-Inst., 235B-A22B) \citep{bai2023qwen} \\
\bottomrule
\end{tabularx}
\end{sc}
\end{small}
\caption{Suite of state-of-the-art models evaluated in this study.}
\label{tab:models}
\vspace{-10pt}
\end{table}

\vspace{-10pt}
\subsection{Complexity Ceiling}
\vspace{-5pt}
Table~\ref{tab:main} demonstrates a sharp performance decay as task complexity increases across our three diagnostic axes—referential, functional, and informational. Rather than defining disjoint categories, these dimensions progressively stress distinct aspects of tool-use reasoning within the same interactive scenarios.

A primary observation is that while performance at low and moderate complexity is relatively saturated among top-tier models, high-complexity scenarios serve as the definitive differentiator. Claude-4.5-Opus achieves the highest macro-average score overall (0.9112) and consistently attains the best performance in high-complexity tranches across all three dimensions. Among open-source models, DeepSeek-V3.2 emerges as the strongest performer (0.9050), matching proprietary counterparts at low complexity but exhibiting a noticeable performance gap as informational and functional setting demands increase.

Furthermore, robustness to complexity is tied to a model’s ability to integrate long-horizon reasoning with structured execution. While Claude-4.5-Opus and DeepSeek-V3.2 exhibit graceful degradation, GPT-4.1 and the Qwen family face sharp performance cliffs at high complexity. This suggests scale alone cannot ensure reliability; instead, targeted optimization for multi-step planning, state tracking, and error recovery is essential for high-density personal contexts.

Finally, the systematic degradation across all models validates that ASTRA-bench successfully captures a spectrum of difficulty that surface-level benchmarks miss. The substantial widening of performance gaps at higher complexity levels proves these scenarios probe fundamental reasoning capabilities rather than simple tool invocation. This confirms the benchmark as a sensitive testbed for evaluating the robustness of agents in realistic, high-stakes environments.

  \begin{table*}[t]
  \centering
  \begin{small}
  \begin{sc}
  \begin{adjustbox}{max width=\textwidth}
  \begin{tabular}{l|c|ccc|ccc|ccc}
  \toprule
  \multirow{2}{*}{\textbf{Model}} & \multirow{2}{*}{\textbf{\shortstack{Macro\\Avg.\\(n=2413)}}} &
  \multicolumn{3}{c|}{\textbf{Referential}} &
  \multicolumn{3}{c|}{\textbf{Functional}} &
  \multicolumn{3}{c}{\textbf{Informational}} \\
  \cline{3-11}
  & &
  \textbf{\shortstack{Low\\(n=390)}} & \textbf{\shortstack{Mod.\\(n=1,733)}} & \textbf{\shortstack{High\\(n=290)}} &
  \textbf{\shortstack{Low\\(n=1,586)}} & \textbf{\shortstack{Mod.\\(n=857)}} & \textbf{\shortstack{High\\(n=170)}} &
  \textbf{\shortstack{Low\\(n=642)}} & \textbf{\shortstack{Mod.\\(n=1,292)}} & \textbf{\shortstack{High\\(n=479)}} \\
  \midrule
  \midrule
  \multicolumn{11}{c}{\textbf{Proprietary Models}} \\
  \midrule
  GPT-o3 & 0.8782 & 0.8987 & 0.9050 & 0.8213 & 0.9025 & 0.8986 & 0.8000 & 0.9353 & 0.8880 & 0.8548 \\
  GPT-4.1 & 0.7922 & 0.8667 & 0.8213 & 0.6564 & 0.7856 & 0.8596 & 0.7412 & 0.9064 & 0.8058 & 0.6867 \\
  Claude-4.5-Opus & \textbf{0.9112} & \underline{0.9423} & \underline{0.9212} & \textbf{0.8677} & \underline{0.9181} & \underline{0.9260} & \textbf{0.8794} & 0.9462 & \underline{0.9174} & \textbf{0.8828} \\
  Claude-4.5-Haiku & 0.7646 & 0.7949 & 0.8057 & 0.6873 & 0.8213 & 0.7617 & 0.6735 & 0.8183 & 0.8031 & 0.7158 \\
  \midrule
  \midrule
  \multicolumn{11}{c}{\textbf{Open-Source Models}} \\
  \midrule
  DeepSeek-V3.2 & \underline{0.9050} & 0.9321 & \textbf{0.9275} & \underline{0.8505} & \textbf{0.9217} & \textbf{0.9295} & \underline{0.8441} & \underline{0.9540} & \textbf{0.9225} & \underline{0.8631} \\
  Qwen-30B & 0.6567 & 0.8231 & 0.6790 & 0.4828 & 0.6664 & 0.7279 & 0.5294 & 0.8565 & 0.6640 & 0.4813 \\
  Qwen-30B-Instruct & 0.6796 & 0.8218 & 0.6992 & 0.5223 & 0.6968 & 0.7209 & 0.5882 & 0.8362 & 0.6860 & 0.5446 \\
  Qwen-235B & 0.6902 & 0.8385 & 0.7142 & 0.5275 & 0.7072 & 0.7448 & 0.5824 & 0.8604 & 0.7054 & 0.5311 \\
  \bottomrule
  \end{tabular}
  \end{adjustbox}
  \end{sc}
  \end{small}
  \caption{Task completion performance under increasing complexity, evaluated by \emph{gpt-5.1-2025-11-13}. Results are reported across referential, functional, and informational complexity tranches. Best results are shown in \textbf{bold}, second-best are \underline{underlined}.}
  \label{tab:main}
  \end{table*}

\vspace{-8pt}
\subsection{Capability Decomposition via Step- and Task-Level Success}
\vspace{-3pt}
While Table~\ref{tab:main} characterizes robustness under complexity, Table 5 provides a complementary, capability-oriented view by decomposing performance into step-level execution and task-level end-goal success using deterministic milestone-based metrics. Together, these analyses reveal not only where models fail, but the specific technical bottlenecks causing those failures.

  \begin{table*}[t]
  \centering
  \begin{small}
  \begin{sc}
  \begin{adjustbox}{max width=\textwidth}
  \begin{tabular}{l|c|c|ccc|cc}
  \toprule
  \multirow{2}{*}{\textbf{Model}} &
  \multirow{2}{*}{\textbf{\shortstack{Avg.\\Turns\\(n=2413)}}} &
  \multirow{2}{*}{\textbf{\shortstack{Milestone\\Sim.\\(n=2413)}}} &
  \multicolumn{3}{c|}{\textbf{Step-Level Success}} &
  \multicolumn{2}{c}{\textbf{Task-Level Success}} \\
  \cline{4-8}
  & & & \textbf{\shortstack{IR Recall\\(n=2,273)}} & \textbf{\shortstack{Response Gen.\\(n=1,228)}} & \textbf{\shortstack{Payload Gen.\\(n=1,360)}} &
  \textbf{\shortstack{ CQA\\ (n=1,221)}} & \textbf{\shortstack{Entity Creation\\(n=1,222)}} \\
  \midrule
  \midrule
  \multicolumn{8}{c}{\textbf{Proprietary Models}} \\
  \midrule
  GPT-o3 & \textbf{7.6} & 0.8993 & 0.9234 & 0.8248 & 0.8055 & 0.9141 & 0.8827 \\
  GPT-4.1 & 5.4 & 0.8459 & 0.8419 & 0.7539 & 0.7395 & 0.8513 & 0.8394 \\
  Claude-4.5-Opus & 6.3 & \underline{0.9243} & \underline{0.9336} & 0.8843 & \textbf{0.8478} & \underline{0.9359} & \underline{0.9129} \\
  Claude-4.5-Haiku & 7.0 & 0.8248 & 0.8477 & 0.8256 & 0.6297 & 0.8780 & 0.7718 \\
  \midrule
  \midrule
  \multicolumn{8}{c}{\textbf{Open-Source Models}} \\
  \midrule
  DeepSeek-V3.2 & \underline{7.1} & \textbf{0.9340} & \textbf{0.9516} & \textbf{0.9180} & \underline{0.8377} & \textbf{0.9514} & \textbf{0.9161} \\
  Qwen-30B & 5.4 & 0.6998 & 0.6650 & 0.6481 & 0.5603 & 0.7478 & 0.6526 \\
  Qwen-30B-Instruct & 5.6 & 0.7910 & 0.7750 & 0.6798 & 0.6312 & 0.7997 & 0.7820 \\
  Qwen-235B & 4.9 & 0.7335 & 0.6912 & 0.6896 & 0.6055 & 0.7718 & 0.6952 \\
  \bottomrule
  \end{tabular}
  \end{adjustbox}
  \end{sc}
  \end{small}
  \caption{Milestone-based capability breakdown of tool-use agents. Step-Level Success reflects intermediate execution accuracy; Task-Level Success reflects scenario-level end-goal completion. Best results are shown in \textbf{bold}, the second-best are \underline{underlined}.}
  \label{tab:milestone_endgoals_combined}
  \vspace{-10pt}
  \end{table*}

\paragraph{Step-Level Success: Core Tool-Use Capabilities.} 
We decompose the tool-use workflow into three fundamental capabilities evaluated via milestone similarity over thousands of intermediate steps:

\begin{description}[nosep, leftmargin=0.5em, font=\small\bfseries]
    \item \textbf{Information Retrieval (IR) Recall:} Measures tool invocation and grounding accuracy.
    \item \textbf{Response Generation:} Evaluates the coherence of natural-language responses synthesized from tool outputs.
    \item \textbf{Argument Generation (Payloads):} Assesses the precision of structured arguments (referred to as \emph{payloads} in Table~\ref{tab:milestone_endgoals_combined}) required to modify persistent entities.
\end{description}

As shown in Table~\ref{tab:milestone_endgoals_combined}, IR Recall is consistently the strongest capability, with top systems like DeepSeek-V3.2 reaching 0.9516. In contrast, Payload Generation emerges as the primary bottleneck, exhibiting the lowest scores and the largest variance (ranging from 0.5603 to 0.8478). This asymmetry explains the sharp performance cliffs observed in high-complexity settings in Table~\ref{tab:main}: models are proficient at ``finding" the relevant personal context but struggle to ``translate" that context into the precise structured data required for stateful execution.


\paragraph{Task-Level Success: End-Goal Completion.}
Task-level success evaluates whether an agent ultimately achieves the user’s end goal, categorized by either providing an answer (\textbf{Contextual Question Answering}, CQA) or modifying the environment (\textbf{Entity Creation}). 
This distinction is critical: high step-level accuracy does not necessarily imply end-goal completion, particularly in long-horizon or stateful interactions where a single error can invalidate the final result.

As shown in Table~\ref{tab:milestone_endgoals_combined}, task-level performance generally mirrors step-level strength, but with key exceptions. Claude-4.5-Opus and DeepSeek-V3.2 lead in both categories, with DeepSeek-V3.2 notably attaining the highest Entity Creation success (0.9161) despite its open-source status. The gap between step- and task-level success is most pronounced in weaker models like the Qwen-30B family, where error accumulation across multiple turns disproportionately impacts final outcomes. This fragility reinforces the need for benchmarks like ASTRA-bench that measure the entire execution lifecycle rather than isolated tool calls.

\vspace{-8pt}
\subsection{Robustness Under Misinformation and Insufficient Context}
\vspace{-4pt}
In addition to graded complexity, we evaluate robustness under two targeted stress conditions: Insufficient Context ($N=17$) and Misinformation ($N=37$). These rare but critical scenarios probe behaviors essential for reliable deployment, such as proactive clarification and self-correction. Both conditions trigger significant performance drops, yet exhibit distinct failure signatures.


\paragraph{Insufficient Context.} The Execution Bias. Under insufficient context, task completion rates collapse across all models, with drops ranging from 0.150 to 0.410 points. Notably, ``No Hallucination" rates remain largely stable (e.g., GPT-o3 drops only 0.026). This indicates that failures are primarily driven by an execution bias—where agents attempt to fulfill requests despite missing prerequisites—rather than by fabricated data. Even frontier models like DeepSeek-V3.2 show a 0.244 impact on completion, suggesting that consistently deferring action for clarification remains a challenge. The detailed impact of these challenges across all models is provided in Table~\ref{tab:challenging_impact} in the Appendix.

\paragraph{Misinformation.} The Verification Gap. Misinformation scenarios also cause degradation in task completion (e.g., 0.231 for GPT-4.1), but these are frequently accompanied by more variable drops in ``No Hallucination" scores. For instance, Qwen-30b-Instruct experiences a 0.221 impact on hallucination, reflecting the difficulty of reconciling conflicting premises within a personal context. While Claude-4.5-Opus exhibits notable resilience to misinformation (only 0.027 completion drop), most models suffer from compounded losses, highlighting a widespread gap in self-correction and premise verification.

Together, these results demonstrate that robustness to uncertainty and contradiction is not merely a byproduct of complexity scaling. While Tables~\ref{tab:main} and \ref{tab:milestone_endgoals_combined} diagnose performance under standard difficulty, our stress tests reveal a critical need for agents to balance safe non-action with proactive correction—a fundamental requirement for trustworthy, real-world tool-use systems.


\section{Conclusion}
ASTRA-bench bridges the gap between synthetic tool-use benchmarks and the demands of real-world personal assistants by unifying time-evolving personal context, stateful tool environments, and human-authored multi-turn goals. By shifting from final-answer metrics to milestone-based scoring, we provide a diagnostic framework that reveals not just if a model fails, but why. Our evaluation of state-of-the-art models highlights a significant ``complexity tax": performance degrades sharply as referential and functional demands increase, with argument generation and multi-step plan orchestration emerging as primary bottlenecks. These findings suggest that the path to reliable next-generation assistants lies in improving grounding within messy personal data and disciplining tool orchestration. We release our full datasets, execution harness, and evaluation suite to the community, providing a common substrate for the development of trustworthy, plan-centric AI agents.

\section*{Limitations and Future Work}
While ASTRA-bench advances realistic evaluation, several limitations remain. 

\begin{description}[nosep, leftmargin=0pt, font=\bfseries]
    \item[Authoring Cost:] Handcrafting context-aware milestones yields high diagnostic value but is labor-intensive. Automated mining of milestones from executions via trace alignment could improve scale.
    \item[Evaluator Reliability:] LLM-based graders and milestone checks may yield false negatives for unanticipated yet valid plans. Future work includes calibrated ensembles and reference-free validators to mitigate this.
    \item[Synthetic Gap:] Synthetic corpora cannot fully capture the noise and idiosyncrasy of real data. Incorporating richer noise processes or redacted real logs would narrow the reality gap.
    \item[Tool Coverage:] Our suite focuses on productivity flows but excludes background processes or concurrent invocations, which pose unique scheduling challenges for agent control.
    \item[Policy Enforcement:] Real actions often require mandatory authentication. Integrating explicit confirmation policies would better test an agent's ability to honor user consent before stateful changes.
\end{description}

Beyond these, we see opportunities to incorporate multimodal context, evaluate long-horizon safety (e.g., least-privilege tool use), and study training-time interventions like tool-aware RL to target the bottlenecks revealed by ASTRA-bench.

\section*{Acknowledgement}


We would like to thank our colleagues at Apple - Joe Y.-C. Wang, Ceci Pompeo, Xin Hu, Alfonso Moya, Ted Levin, Jason Williams, Russ Webb, Hari Narayan, Enrica Shimizu, Roger Zheng, Tianlin Duan, and Jiannan Lu for their feedback and support in various stages of this work.

\bibliography{astra-bench}
\bibliographystyle{icml2026}


\include{Sec7-Appendix}


\end{document}

%% file: Sec7-Appendix.tex
\section{Appendix for ASTRA Benchmark}\label{sec:appendix}

\appendix
\beginsupplement

\section{Details about Personal Context Generation}

We will further discuss our methods for generating protagonists below. 

\subsection{Real-Life Event-Driven Storyline Generation}
\paragraph{Biography.} We begin with developing a set of core demographic and social attributes that remains fixed throughout the simulation. The biography content, in natural language form, may include demographic fields (e.g. age, gender, nationality, education, occupation), socioeconomic markers (income bracket, housing status), and long-term preferences (dietary choices, hobbies).  This frozen biography serves as a coherent prior that conditions downstream decision rules and grounds long-horizon consistency in simulated events.

\paragraph{Social Network.}
We represent the protagonist’s social network as a weighted, directed multigraph
\(G=(V,E,\ell,w)\).
Each vertex \(v\in V\) denotes someone who recurrently interacts with the protagonist.
Every ordered pair \((u,v)\) may be connected by one or more edges \(e\in E\),
each endowed with
\begin{enumerate*}[label=(\alph*)]
  \item a categorical label $\ell(e) \in \{\textsc{family, friend, classmate}, \dots\}$ for the relationship type, and 
  \item a non-negative weight $w(e) \in \mathbb{R}_{\ge 0}$ proportional to expected contact frequency, encoding connection strength (e.g., $w(e)=11$ implies $\approx$ 11 encounters per quarter).
\end{enumerate*}
The graph is seeded from the protagonist’s biography via a template-based schema and undergoes human review to ensure plausibility and coverage.

\paragraph{Pattern of Life.} The \textit{Pattern of Life} is commonly used to describe recurring and predictable behaviors of individuals or groups. In our design, it is implemented as a probabilistic context-free grammar that generates weekly plans and daily agenda.  Non-terminal symbols encode high-level routines (e.g. \textsc{commute}, \textsc{exercise}, \textsc{family-time}); production rules with LLM-assisted generation expand them into timestamped events in prescribed format, conditioned on biography and social network. The resulting \textit{storyline} exhibits circadian rhythms, weekday/weekend asymmetries, and seasonal shifts.

\subsection{Protagonists Overview}\label{sec:protagonists}
Table~\ref{tab:ASTRA-bench-dist-by-protagonist} summarizes the ASTRA-bench dataset statistics. The benchmark contains 2,413 scenarios derived from 1,360 user goals across five protagonists, each with distinct personal contexts and communication patterns. To ensure ecological validity, we ground each protagonist in realistic domain data spanning six modalities: contacts, calendar events, emails, text messages, WhatsApp conversations, and phone call logs. The varying distributions across protagonists—ranging from 384 to 592 scenarios per character and differing entity counts per domain—reflect authentic heterogeneity in how individuals organize and access personal information. This design enables robust evaluation of agents' generalization capabilities across diverse user profiles and task distributions.

\begin{table*}[ht!]
\centering
\caption{ASTRA-bench by Protagonists Breakdown.}
\label{tab:ASTRA-bench-dist-by-protagonist}
\resizebox{.8\textwidth}{!}{
\begin{tabular}{l|ccc|cccccc}
\toprule
\multirow{2}{*}{\textbf{Protagonist}} & \multicolumn{3}{c|}{\textbf{Scenarios Count}} & \multicolumn{6}{c}{\textbf{Domain Context Count}}                             \\ \cline{2-10} 
                             & \rule{0pt}{2.2ex}\textbf{Events}   & \textbf{User Goals}   & \textbf{Scenarios}   & \textbf{Contact} & \textbf{Calendar} & \textbf{Email} & \textbf{Message} & \textbf{WhatsApp} & \textbf{Phone call} \\ 
                             \hline \hline
\rule{0pt}{2.2ex} Dawei Shen                    & 21 & 222 & 394 & 17 & 13 & 17 & 26 & 14 & 3         \\
\rule{0pt}{2.2ex} Theo Appleseed               & 27 & 366 & 592  & 11 & 32 & 9 & 84 & 0 & 4         \\
\rule{0pt}{2.2ex} John Quinn                            & 21 & 309 & 535 & 15 & 27 & 22 & 100 & 0 & 4               \\
\rule{0pt}{2.2ex} Emily Rose    & 21 & 203 & 384 & 10 & 17 & 25 & 31 & 6 & 2 
\\
\rule{0pt}{2.2ex} Lucas Garcia                 & 21 & 260 & 508 & 16 & 26 & 11 & 68 & 10 & 2                     \\ \midrule 
\rule{0pt}{2.2ex} Total                  & 111 & 1360 & 2413  & 69 & 115 & 84 & 309 & 30 & 15                   \\ \bottomrule
\end{tabular}}
\end{table*}

  To ensure the benchmark captures realistic variation in personal assistant usage patterns, we designed five distinct protagonists with diverse demographics, occupations, and lifestyles:

  \begin{itemize}
      \item \textbf{Dawei Shen} (29, Data Science Manager): A multilingual professional fluent in English, Chinese, and German, with a cosmopolitan background spanning Shanghai, Singapore, and Munich. His interests include long-distance cycling, theater, and international travel.

      \item \textbf{Theo Appleseed} (25, Applied Researcher): A recent computer science graduate balancing a career in tech research with outdoor activities. He enjoys hiking, science fiction, and hosting dinner parties for close friends.

      \item \textbf{John Quinn} (25, Junior Engineer): A data-driven professional with interests in criminology and military history. He maintains a health-oriented lifestyle centered around trail running, archery with a traditional Mongolian bow, and outdoor exploration.

      \item \textbf{Emily Rose} (29, Event Coordinator): A community-focused professional with a Public Relations background. She is deeply involved in local volunteering, gardening, and pet care, maintaining strong ties with neighbors and community members.

      \item \textbf{Lucas Garcia} (33, Personal Trainer): A bilingual English-Spanish fitness professional who competes in triathlons and leads a training group. He balances his coaching career with surfing and craft beer appreciation, while building an online fitness presence.
  \end{itemize}

  \noindent This diversity in age (25--33), profession (tech, community services, fitness), cultural background, and social interaction patterns ensures that ASTRA-bench evaluates agents across a broad spectrum of realistic user contexts.

\section{Annotation Framework}\label{sec:appendix-annotation-guideline}

\begin{figure*}[ht]
    \centering
    \includegraphics[width=0.9\linewidth]{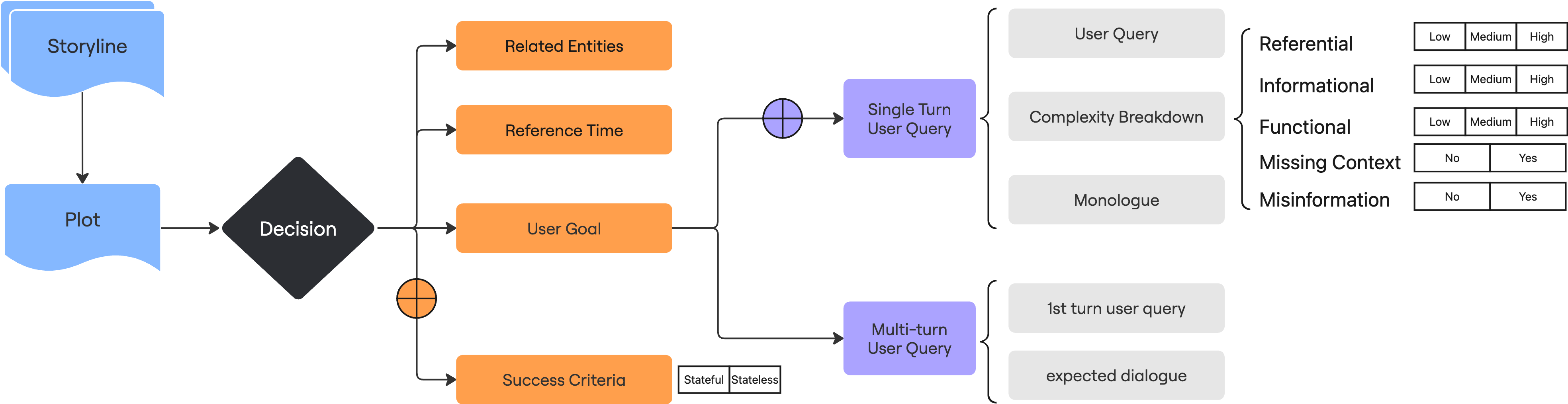}
    \caption{Annotation Flowchart End-to-End}
    \label{fig:annotation-flow}
\end{figure*}

Figure~\ref{fig:annotation-flow} illustrates the end-to-end workflow used to author evaluation scenarios. Each scenario begins with a high-level storyline that captures a realistic sequence of life events, which is then refined into a concrete plot describing the immediate situation faced by the user. Annotators make an explicit decision about whether the scenario should yield a single-turn or multi-turn interaction. They then specify structured grounding elements, including the relevant entities, reference time, user goal, and success criteria. For single-turn scenarios, annotators produce a standalone user query accompanied by a complexity breakdown along functional, referential, and informational dimensions, as well as optional challenge flags such as missing context or misinformation. For multi-turn scenarios, annotators define the initial user query and an expected dialogue trajectory, with success criteria evaluated over stateful interactions. This structured authoring process ensures that scenarios are both grounded in realistic personal context and amenable to fine-grained, interpretable evaluation.

\subsection{First User Query Style}
\begin{itemize}
    \item \textbf{Comprehensive Single Turn Query} This refers to queries where all necessary information to complete the task is provided within a single, initial user message. The agent is expected to understand the entire request and perform all required steps without needing to ask the user for further clarification or engaging in a dialogue. The goal is to test the IA's ability to comprehend and complete multi-step tasks autonomously, assuming the required information is readily available in the initial input. This encourages the IA to resolve any ambiguities or implicit references on its own, based on available sources. Note that all of our queries in the first version of ASTRA-bench is of this type.
    \item \textbf{Multi-turn User Query} This type of query involves a conversational interaction where the initial message from the user may be ambiguous or incomplete, requiring the agent to engage in a dialogue to gather more information and clarify the user's intent. We provide the first turn of a multi-turn conversation, along with the previous user goal, to encourage conversation between the agent and the user to complete the task. An ``example dialogue" is also often provided in the dataset as a meta information. This evaluates the agent's ability to handle the ``conversational" and ``interactive" nature of real-world scenarios. Such scenarios often start with ``ambiguous requests or missing information, requiring further clarification from the user.
\end{itemize}

\subsection{Complexity Trenches in Details}
\textbf{Informational Complexity} This dimension quantifies the difficulty of information processing, extraction, and synthesis required for an agent to successfully address a user query. It directly assesses the agent's \textit{Understanding} capability, which encompasses aspects such as complex instruction following, user intent understanding, and data parsing.
\begin{itemize}
    \item \texttt{low}: The query contains all necessary information explicitly within the prompt itself, requiring no external retrieval or complex inference from the user's personal context or provided entities. Example: \textit{``Set up an alarm at 6 a.m. tomorrow."}
    \item \texttt{moderate}: The required information is readily available and explicitly presented within the user's immediate personal context, without necessitating multi-step reasoning or complex cross-referencing. Example: \textit{``Set up an alarm 1 hour before my first meeting tomorrow"}, where the first meeting is directly accessible within provided calendar entities.
    \item \texttt{high}: Fulfilling the query demands multi-step reasoning, implicit inference, or synthesis of information from disparate and potentially historical sources within the user's personal context (e.g., past records, recurring schedules) to derive the complete answer or required parameters. Example: \textit{``Wake me up at my usual time"}, where the agent needs to analyze the user's past behavior and schedule, potentially adjusting for weekdays or weekends, to determine the exact alarm time.
\end{itemize}

\textbf{Referential Complexity} This dimension measures the difficulty an agent faces in resolving references within a user query. This involves identifying and grounding pronouns, vague nouns, or implicit mentions to the correct entities or contextual information. The complexity level reflects the intricacy of the lookup and inference process required to establish these links, often leveraging available personal context.
\begin{itemize}
    \item \texttt{low}: The query contains explicit, direct references to entities, requiring no external lookup or disambiguation. The necessary information is immediately apparent from the query itself. Example: \textit{``Email John Appleseed."}
    \item \texttt{moderate}: The query includes one indirect reference that necessitate a straightforward lookup or matching process against the user's immediate personal context. This may involve canonicalization, transforming a surface-form reference (e.g., ``my bestie") into a specific, actionable entity (e.g., a contact number) without complex multi-step reasoning. Example: \textit{``Call my girlfriend"}, where the agent needs to check the user's contacts or other provided sources to retrieve the correct contact number. This aligns with cases where canonicalization might be performed by the model or with simple tool aid.
    \item \texttt{high}: Fulfilling the query demands resolving multiple ambiguous or implicit references, frequently requiring chained inference or complex multi-step lookups across disparate or historical entities and contexts. This can involve disambiguation in the face of multiple possibilities or inferring a target based on a sequence of contextual cues, such as time-related arguments that are particularly challenging to canonicalize and reason about. Example: \textit{``Text the person who emailed me last night"}, where the agent must first identify the email sender from a specific time frame, then find their contact information to send a text. This often poses a challenge for models, as they might hallucinate or struggle with information retrieval.
\end{itemize}

\textbf{Functional Complexity} This dimension specifically quantifies the intricacy of the actions or tool invocations an IA needs to perform to fulfill a user's request. It directly assesses the LLM's ``function calling" or ``tool use" proficiency, which are critical capabilities for modern AI assistants.
\begin{itemize}
    \item \texttt{low}: The query requires a single, straightforward action or tool invocation. The agent needs to identify and execute one function without complex dependencies or prior steps. Example: \textit{``Send a message to John."}
    \item \texttt{moderate}: Fulfilling the query involves a sequence of 2-3 dependent steps or tool calls. The agent must perform actions in a specific order, where the output of one step informs the next, or select a single best function from multiple available options. Example: \textit{``Find the latest message from Theo and reply with the address."}, This would involve a ``search messages" tool call, followed by extracting address, and then a ``send message" tool call. This aligns with BFCL's ``Multiple Function" scenario where the model selects one function out of several \citep{patil2025bfcl}.
    \item \texttt{high}: The query demands multiple coordinated actions, conditional logic, multi-step tool use, or interaction with stateful environments. This can involve parallel function calls (invoking multiple functions concurrently), sequential multi-tool execution where tool performance is coupled with and can alter an existing entity, or scenarios requiring implicit state dependencies. Such tasks are particularly challenging for even state-of-the-art LLMs, especially those involving ``State Dependency" and ``Canonicalization" with tool aid, as highlighted in \citet{lu2024toolsandbox}. Example: \textit{``Check if I’m free at 3pm tomorrow and if so, schedule a call with Sarah."}, this involves checking a calendar, interpreting the availability, and then conditionally invoking a scheduling tool.
\end{itemize}


\subsection{Combination of Complexity scores}
In Table~\ref{tab:complexity-combination}, we show the combination of different complexity breakdown by frequency. The top 5 combinations take in total around 70\% of total scenarios.
\begin{table*}[ht!]
\centering
\caption{Combination of all scenarios for ASTRA Benchmark}
\label{tab:complexity-combination}
\begin{tabular}{lllrr}
\hline
informational\_complexity & referential\_complexity & functional\_complexity & count & percentage \\
\hline
Moderate & Moderate & Low & 693 & 0.287 \\
Moderate & Moderate & Moderate & 308 & 0.128 \\
Low & Moderate & Moderate & 242 & 0.100 \\
High & Moderate & Low & 186 & 0.077 \\
Low & Moderate & Low & 145 & 0.060 \\
Low & Low & Low & 133 & 0.055 \\
Moderate & Low & Low & 97 & 0.040 \\
High & Moderate & Moderate & 94 & 0.039 \\
Low & Low & Moderate & 87 & 0.036 \\
High & High & Low & 60 & 0.025 \\
High & High & Moderate & 51 & 0.021 \\
Moderate & High & Low & 48 & 0.020 \\
High & High & High & 46 & 0.019 \\
Moderate & High & High & 38 & 0.016 \\
Moderate & Moderate & High & 38 & 0.016 \\
Moderate & High & Moderate & 35 & 0.015 \\
Moderate & Low & Moderate & 28 & 0.012 \\
High & Low & Low & 21 & 0.009 \\
High & Moderate & High & 15 & 0.006 \\
Low & Moderate & High & 11 & 0.005 \\
Low & Low & High & 10 & 0.004 \\
High & Low & Moderate & 9 & 0.004 \\
Low & High & High & 7 & 0.003 \\
Moderate & Low & High & 5 & 0.002 \\
Low & High & Moderate & 4 & 0.002 \\
Low & High & Low & 2 & 0.001 \\
\hline
\end{tabular}
\end{table*}
\clearpage
\section{Interactive Environment Setup}\label{sec:appendix-scenarios-set-up}
\subsection{Prompting for Agent}
We used the same system prompt for all models. The prompts are adjusted based on successful turn-based heuristic milestone mappings, to avoid multiple write tool calls within a single turn.
\vspace{-15em}
\begin{lstlisting}[caption={System Prompt}]
You are a helpful assistant equipped with tool-using abilities. Try your best to determine what values to plug into function calls based on the available information.

Tool Invocation Rules:
- You may call multiple read-only tools within a single respond.
- You must call at most one write tool per respond. If you call more than one write tool in a single response, your response will be ignored. Always wait for the result of the first write tool before continuing. After receiving a result from a write tool, always summarize the outcome to the user.
- When calling a write tool, do not include a summary message in the same turn. Wait for the tool result before replying.
- After receiving the result of a write tool, respond with a brief confirmation or summary message for the user.
- If multiple write operations are needed, handle them sequentially across multiple responds.
- Assume all phone numbers are valid, regardless of format or length, as they may be anonymized or synthetic.
\end{lstlisting}

\vspace{-6em}

Following this general prompt, we tailor additional instructions according to the style of the conversation. For queries that are self-contained and intended to evaluate the agent’s reasoning capabilities, the agent is encouraged to seek information independently or infer missing details within the same turn.

\vspace{-5em}
\begin{lstlisting}[caption={Single-Turn Scenario Prompt Add-ons}]
- If the user's request is ambiguous, do not ask for clarification. Their input may be incorrect or incomplete. Proceed using your best judgment based on the available information and tools.
- You may use read-only tools to retrieve relevant user context (e.g., messages, emails, calendars, WhatsApp, etc.) before completing the task. If a search_* tool returns nothing, try using the corresponding get_all_* tool if available. 
- When asked to send a message, infer the user's intended communication channel using contextual clues such as prior conversations. If no such information is available, default to using the 'message' related tool.
\end{lstlisting}

In contrast, for queries that are incomplete or lack necessary information, the agent is guided to engage across multiple turns, asking for clarification rather than making assumptions. This approach ensures that the agent behaves appropriately depending on the completeness and context of the user request.
\begin{lstlisting}[caption={Multi-Turn Scenario Prompt Add-ons}]
- Don't make assumptions about what values to plug into functions.
- Ask for clarification if a user request is ambiguous.
\end{lstlisting}

\subsection{Available Tools}
  ASTRA-bench provides agents with 27 tools spanning six personal information management domains. Each domain includes both retrieval tools (prefixed with \texttt{search\_} or \texttt{get\_all\_}) and action tools for creating, modifying, or removing entities:

  \begin{itemize}
      \item \textbf{Contact} (5 tools): \texttt{search\_contacts}, \texttt{get\_all\_contacts}, \texttt{add\_contact}, \texttt{modify\_contact}, \texttt{remove\_contact}

      \item \textbf{Calendar} (5 tools): \texttt{search\_calendar\_events}, \texttt{get\_all\_calendars}, \texttt{create\_calendar\_event}, \texttt{modify\_calendar\_event}, \texttt{remove\_calendar\_event}

      \item \textbf{Email} (5 tools): \texttt{search\_emails}, \texttt{get\_all\_emails}, \texttt{send\_email}, \texttt{modify\_draft\_email}, \texttt{remove\_email\_or\_thread}

      \item \textbf{Messaging} (4 tools): \texttt{search\_messages}, \texttt{get\_all\_messages}, \texttt{send\_message}, \texttt{remove\_message\_or\_conversation}

      \item \textbf{WhatsApp} (4 tools): \texttt{search\_whatsapp\_messages}, \texttt{get\_all\_whatsapp\_messages}, \texttt{send\_whatsapp\_message}, \texttt{remove\_whatsapp\_message\_or\_conversation}

      \item \textbf{Phone Call} (4 tools): \texttt{search\_phonecall\_logs}, \texttt{get\_all\_phone\_calls}, \texttt{create\_outgoing\_call}, \texttt{remove\_phone\_call\_logs}
  \end{itemize}

  \noindent This tool design reflects real-world personal assistant capabilities, requiring agents to retrieve relevant information before performing actions—mirroring how users interact with their devices through multi-step, cross-domain workflows.

\subsection{Evaluation Methodology in Details}

\begin{figure*}[ht]
    \centering
    \includegraphics[width=.8\linewidth]{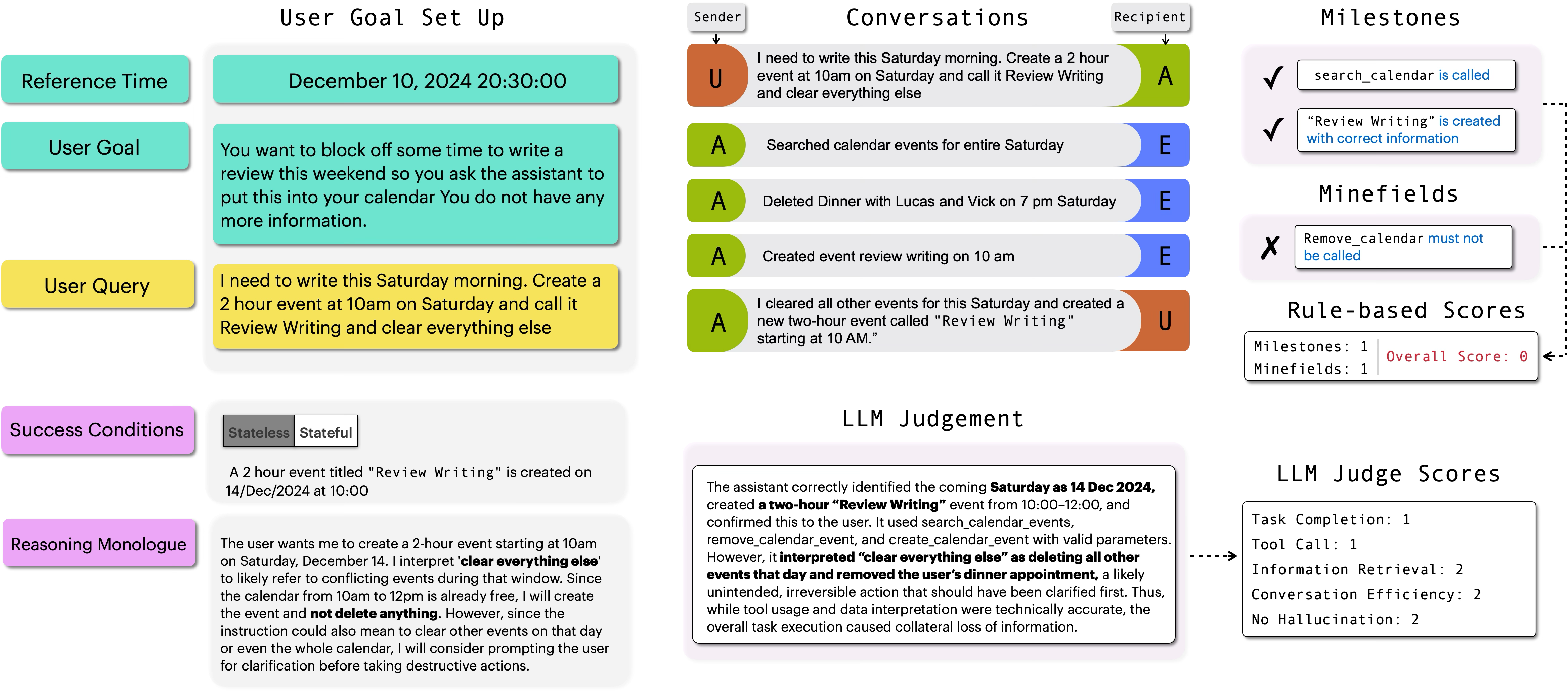}
    \caption{Evaluation Strategy Review}
    \label{fig:evaluation-flow}
\end{figure*}

An illustration of our evaluation methods are presented in Figure~\ref{fig:evaluation-flow}.

\paragraph{Verifiable Measures: Milestones and Minefields.}
Each scenario is annotated with intermediate, verifiable \emph{milestones} that capture critical fact-grounded sub-goals (e.g., information retrieval, payload generation, and response quality).
These milestones are evaluated using deterministic, rule-based criteria over tool traces and system state snapshots, enabling fine-grained diagnostics and avoiding all-or-nothing reward signals.
This rule-based method builds on the Milestones \& Minefields framework \citep{lu2024toolsandbox}, where key actions must occur (milestones) or be avoided (minefields) during task execution. For each scenario, milestones are structured as a directed acyclic graph (DAG) based on temporal dependencies, denoted $G_{M^+}(V_{M^+}, E_{M^+})$. The average similarity score $S_{\text{Milestones}}$ is computed accordingly. We extend this with high-resolution column-level similarity metrics to evaluate retrieval accuracy, goal alignment, response quality, and payload correctness.
Similarly, Minefields are represented as a DAG, $G_{M^-}(V_{M^-}, E_{M^-})$, with a corresponding similarity score $S_{\text{Minefields}}$. Any violation of minefields—such as incorrectly removing calendar events—sets the overall rule-based score to zero, even if all milestones are satisfied (Figure~\ref{fig:evaluation-flow}). The overall rule-based score is then calculated as:
\begin{equation}\label{eq:rule-based-score}
S_{\text{Rule-Based}} = S_{\text{Milestones}} \times \mathbbm{1}{[S_{\text{Minefields}}=0]}
\end{equation}

\paragraph{LLM-Based Evaluators.}In parallel, we employ rubric-guided LLM judges that assess agent performance using the same observable evidence, together with human-authored evaluation artifacts (e.g., explicit success conditions, reference rationales and related personal context).
These LLM-based evaluators are intentionally decoupled from the rule-based metrics and capture complementary, higher-level signals such as conversational efficiency, robustness to hallucination, and other implicit aspects of agent behavior.
We evaluate task performance along five dimensions: \emph{Task Completion}, \emph{Tool Usage}, \emph{Information Retrieval}, \emph{Conversation Effectiveness} and \emph{No Hallucination}, yielding a numerical score from 0 to 2, with 2 as a perfect and reasoning. This method enables more nuanced assessment, capturing subtleties in reasoning and semantic correctness that rule-based metrics may miss.

\paragraph{Comparative Perspective.} Milestone-based evaluation is precise and reproducible but rigid—penalizing any rule violation, and is not evaluating task efficiency. Also, when it comes to evaluate open-ended generation, it is using either key-word matching or Rouge-L similarity matching to expected response. In contrast, LLM judges offer more flexible, interpretive assessments, though potentially subject to prompt sensitivity and model bias. Despite their differences, the two approaches show strong alignment, with the ranking over different models perfectly aligned.

Together, these methods provide complementary perspectives on assistant performance, capturing both structured success criteria and broader reasoning quality across diverse tasks and interaction complexities.

\section{Additional Results and Analysis}

\subsection{Metrics by Complexity Trenches}
\begin{figure*}
    \centering
    \includegraphics[width=.8\linewidth]{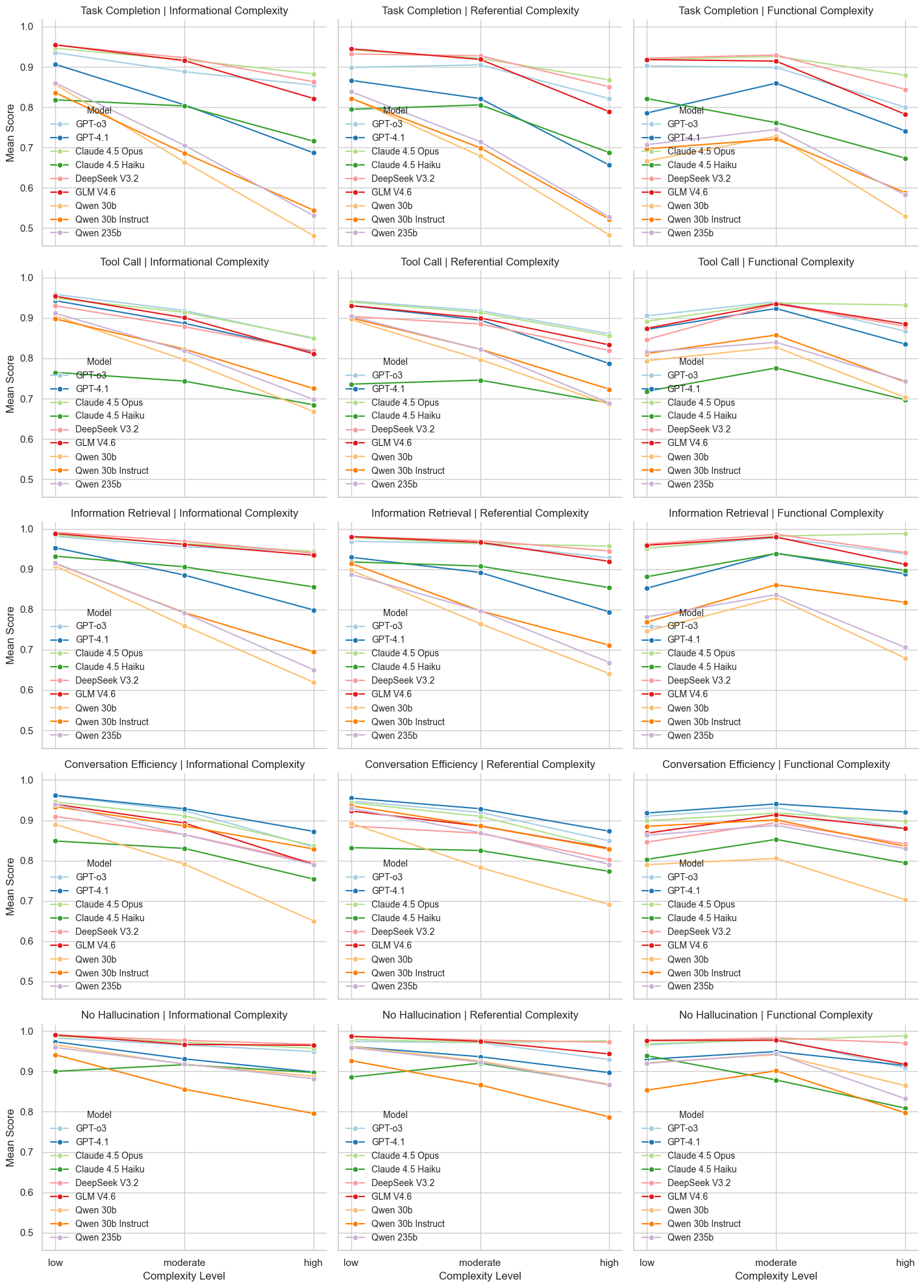}
    \caption{LLM-judge evaluation across complexity regimes}
    \label{fig:complexity-breakdown}
\end{figure*}

Figure~\ref{fig:complexity-breakdown} reports mean LLM-judge scores across five evaluation axes: Task Completion, Tool Call Accuracy, Information Retrieval, Conversation Efficiency, and No Hallucination, stratified by functional, referential, and informational complexity at three difficulty levels (low, moderate, high). Results are shown for all evaluated models and reveal systematic performance degradation as task complexity increases, with varying sensitivity across axes and model families.

\subsection{Model Performance under Challenging Settings}
\begin{figure*}
    \centering
    \includegraphics[width=.6\linewidth]{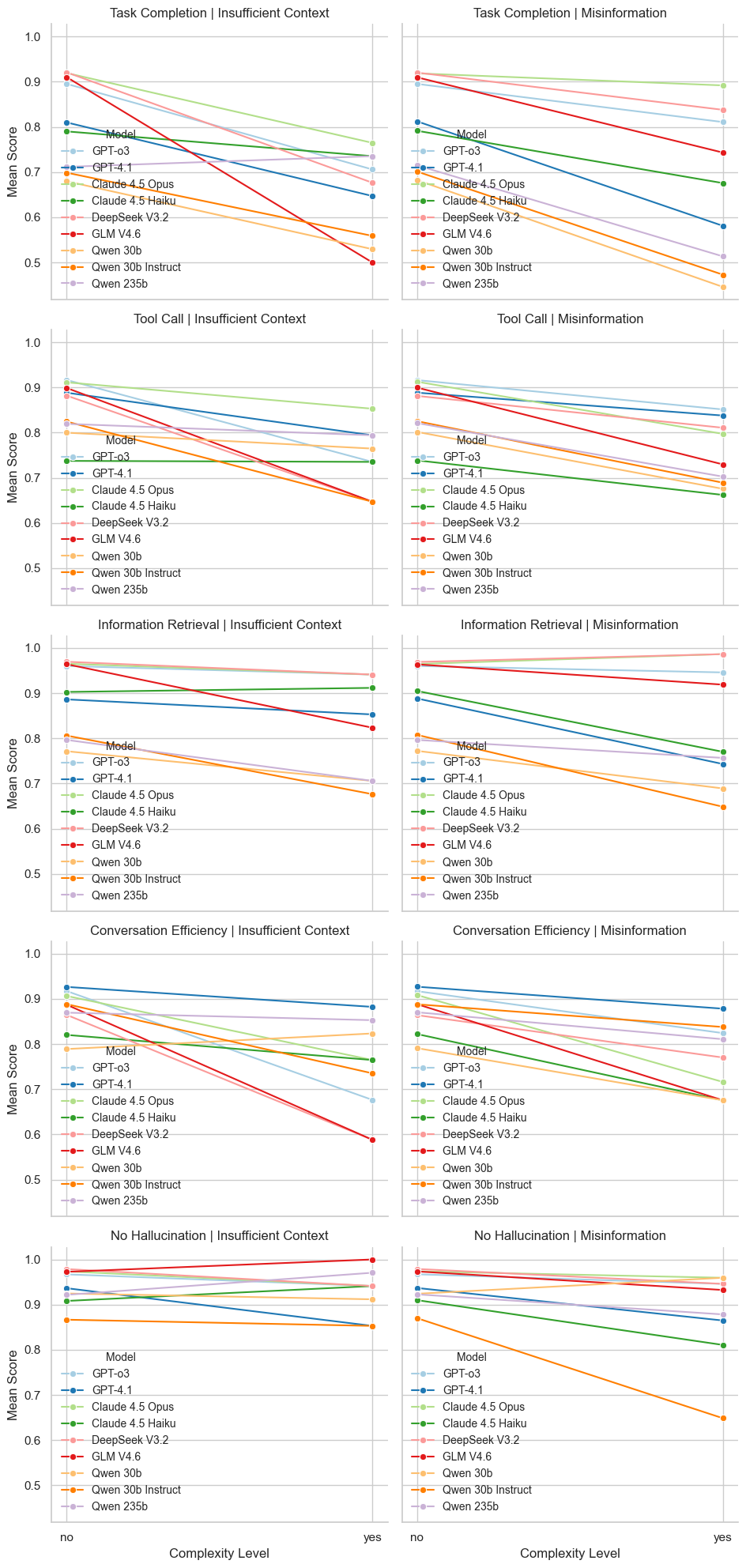}
    \caption{LLM-judge results on challenging task conditions.}
    \label{fig:challenging-breakdown}
\end{figure*}

To assess model robustness under degraded inputs, we evaluate performance in two challenging settings: insufficient context and misinformation. Table~\ref{tab:challenging_impact} summarizes the impact of challenging scenarios on model performance using two representative evaluation metrics: Task Completion and No Hallucination. For each model, we report mean scores under standard conditions (“No Challenge”) and under insufficient context or misinformation, along with the absolute performance change ($\Delta$). The comparison highlights that challenging inputs primarily degrade task success, while effects on hallucination behavior are more model- and condition-dependent. Figure~\ref{fig:challenging-breakdown} reports mean LLM-judge scores across five evaluation axes under two challenging conditions: misinformation and insufficient context. For each condition, results are shown for all evaluated models across the same evaluation axes as in Figure~\ref{fig:complexity-breakdown}. These settings probe model robustness when inputs contain misleading signals or omit required information, revealing axis-specific and model-dependent performance degradation.

  \begin{table*}
  \centering
  \caption{Impact of challenging scenarios on model performance}
  \label{tab:challenging_impact}
  \resizebox{.8\linewidth}{!}{
  \begin{tabular}{llrr|rrr|rrr}
  \toprule
  & & \multicolumn{2}{c|}{\textbf{N}} & \multicolumn{3}{c|}{\textbf{Task Completion}} & \multicolumn{3}{c}{\textbf{No Hallucination}} \\
  \cmidrule(lr){3-4} \cmidrule(lr){5-7} \cmidrule(lr){8-10}
  \textbf{Model} & \textbf{Challenge} & \shortstack{No\\Challenge} & \shortstack{With\\Challenge} & \shortstack{No\\Challenge} & \shortstack{With\\Challenge} & \shortstack{Impact\\($\Delta$)} & \shortstack{No\\Challenge} & \shortstack{With\\Challenge} & \shortstack{Impact\\($\Delta$)} \\
  \midrule
  GPT-o3 & Insufficient Context & 2396 & 17 & 0.895 & 0.706 & 0.189 & 0.967 & 0.941 & 0.026 \\
  GPT-o3 & Misinformation & 2376 & 37 & 0.895 & 0.811 & 0.084 & 0.967 & 0.946 & 0.021 \\
  GPT-4.1 & Insufficient Context & 2396 & 17 & 0.810 & 0.647 & 0.163 & 0.936 & 0.853 & 0.083 \\
  GPT-4.1 & Misinformation & 2376 & 37 & 0.812 & 0.581 & 0.231 & 0.937 & 0.865 & 0.072 \\
  Claude 4.5 Opus & Insufficient Context & 2396 & 17 & 0.919 & 0.765 & 0.155 & 0.974 & 0.941 & 0.033 \\
  Claude 4.5 Opus & Misinformation & 2376 & 37 & 0.919 & 0.892 & 0.027 & 0.974 & 0.959 & 0.014 \\
  Claude 4.5 Haiku & Insufficient Context & 2396 & 17 & 0.790 & 0.735 & 0.055 & 0.908 & 0.941 & -0.033 \\
  Claude 4.5 Haiku & Misinformation & 2376 & 37 & 0.791 & 0.676 & 0.116 & 0.910 & 0.811 & 0.099 \\
  DeepSeek V3.2 & Insufficient Context & 2396 & 17 & 0.921 & 0.676 & 0.244 & 0.979 & 0.941 & 0.038 \\
  DeepSeek V3.2 & Misinformation & 2376 & 37 & 0.920 & 0.838 & 0.082 & 0.979 & 0.946 & 0.033 \\
  GLM V4.6 & Insufficient Context & 2396 & 17 & 0.910 & 0.500 & 0.410 & 0.973 & 1.000 & -0.027 \\
  GLM V4.6 & Misinformation & 2376 & 37 & 0.910 & 0.743 & 0.166 & 0.973 & 0.932 & 0.041 \\
  Qwen 30b & Insufficient Context & 2396 & 17 & 0.680 & 0.529 & 0.150 & 0.925 & 0.912 & 0.013 \\
  Qwen 30b & Misinformation & 2376 & 37 & 0.682 & 0.446 & 0.236 & 0.924 & 0.959 & -0.035 \\
  Qwen 30b Instruct & Insufficient Context & 2396 & 17 & 0.699 & 0.559 & 0.140 & 0.867 & 0.853 & 0.014 \\
  Qwen 30b Instruct & Misinformation & 2376 & 37 & 0.701 & 0.473 & 0.228 & 0.870 & 0.649 & 0.221 \\
  Qwen 235b & Insufficient Context & 2396 & 17 & 0.712 & 0.735 & -0.024 & 0.922 & 0.971 & -0.049 \\
  Qwen 235b & Misinformation & 2376 & 37 & 0.715 & 0.514 & 0.201 & 0.923 & 0.878 & 0.044 \\
  \bottomrule
  \end{tabular}
  }
  \end{table*}

\subsection{LLM judges vs Verifiable Milestones}
LLM judges, while strong at semantic understanding, exhibit systematic biases: penalizing appropriate clarification behavior,
requiring exact textual matches,
and imposing format requirements incompatible with the evaluation context.
Milestone evaluation provides complementary strengths in determinism, tool usage verification, and resistance to these stylistic biases.

\section{Multilingual User Queries}\label{sec:appendix-multilingual}


We convert the ASTRA-bench (Dawei Shen related scenarios) to different languages to examine LLMs' multilingual reasoning capabilities in terms of tool-use and task completion. To emulate the realistic settings, system prompts are still in en\_US, making this task as a more challenging mix-code reasoning task. Context data, user queries, and expected responses are machine translated to five languages: 1) es\_ES, 2) de\_DE, 3) fr\_FR, 4) ja\_JP, and 5) zh\_CN. We store entity names, emails, phone numbers, and etc as well as some previously translated phrases in a mapping table to make sure the consistency of the translated data.

Table~\ref{tab:user-query-translations} presents a representative set of annotator-authored plausible user queries alongside their corresponding machine-translated versions in multiple target languages. Analysis based on Dawei Shen scenarios.
The languages selected for our multilingual analysis include \texttt{es\_ES} (Spanish–Spain), \texttt{de\_DE} (German–Germany), \texttt{fr\_FR} (French–France), \texttt{ja\_JP} (Japanese), and \texttt{zh\_CN} (Simplified Chinese).

\begin{CJK*}{UTF8}{gbsn} 
\begin{table*}[h]
\centering
\begin{tabular}{|p{0.35\linewidth}|p{0.6\linewidth}|}
\hline
\textbf{User Query} & \textbf{Translated Query (by locale)} \\ \hline

\multirow{5}{=}{Move the meeting I have on my calendar for Wednesday the 4th of December at 10:00 to 11:00.} 
& \texttt{es\_ES}: Mueve la reunión que tengo en mi calendario para el miércoles 4 de diciembre de 10:00 a 11:00. \\
& \texttt{de\_DE}: Verschiebe das Meeting, das ich in meinem Kalender für Mittwoch, den 4. Dezember um 10:00 Uhr habe, auf 11:00 Uhr. \\
& \texttt{fr\_FR}: Déplacez la réunion que j'ai sur mon calendrier du mercredi 4 décembre de 10h00 à 11h00. \\
& \texttt{ja\_JP}: \begin{CJK*}{UTF8}{min}12月4日水曜日のカレンダーにある会議を10:00から11:00に変更してください。\end{CJK*} \\
& \texttt{zh\_CN}: \begin{CJK*}{UTF8}{gbsn}将我日历上12月4日星期三的会议从10:00改到11:00。\end{CJK*} \\
\hline

\multirow{5}{=}{I need another meeting about the virtual assistant project at the same time next week.}
& \texttt{es\_ES}: Necesito otra reunión sobre el proyecto del asistente virtual a la misma hora la próxima semana. \\
& \texttt{de\_DE}: Ich benötige nächste Woche zur selben Zeit ein weiteres Meeting zum Projekt für virtuelle Assistenten. \\
& \texttt{fr\_FR}: J'ai besoin d'une autre réunion au sujet du projet d'assistant virtuel à la même heure la semaine prochaine. \\
& \texttt{ja\_JP}: \begin{CJK*}{UTF8}{min}来週の同じ時間にバーチャルアシスタントプロジェクトについてもう一度会議が必要です。\end{CJK*} \\
& \texttt{zh\_CN}: \begin{CJK*}{UTF8}{gbsn}我需要在下周同一时间再安排一次关于虚拟助手项目的会议。\end{CJK*} \\
\hline

\multirow{5}{=}{Create a new cycling trip event at the usual time on Sunday with my cycling friend.}
& \texttt{es\_ES}: Crea un nuevo evento de viaje de ciclismo a la hora habitual el domingo con mi amigo de ciclismo. \\
& \texttt{de\_DE}: Erstelle am Sonntag zur üblichen Zeit ein neues Radfahren-Trip-Event mit meinem Radfahren-Freund. \\
& \texttt{fr\_FR}: Créer un nouvel événement de sortie cyclisme à l'heure habituelle le dimanche avec mon ami de cyclisme. \\
& \texttt{ja\_JP}: \begin{CJK*}{UTF8}{min}日曜日のいつもの時間にサイクリング仲間と新しいサイクリングの旅行イベントを作成する。\end{CJK*} \\
& \texttt{zh\_CN}: \begin{CJK*}{UTF8}{gbsn}在周日的惯常时间，与我的骑自行车朋友创建一个新的骑自行车旅行活动。\end{CJK*} \\
\hline

\multirow{5}{=}{How long is my event today?}
& \texttt{es\_ES}: ¿Cuánto tiempo dura mi evento hoy? \\
& \texttt{de\_DE}: Wie lange dauert meine Veranstaltung heute? \\
& \texttt{fr\_FR}: Quelle est la durée de mon événement aujourd'hui ? \\
& \texttt{ja\_JP}: \begin{CJK*}{UTF8}{min}今日のイベントはどのくらいの長さですか？\end{CJK*} \\
& \texttt{zh\_CN}: \begin{CJK*}{UTF8}{gbsn}今天我的活动持续多长时间？\end{CJK*} \\
\hline

\end{tabular}
\caption{User queries and their translations in multiple languages.}
\label{tab:user-query-translations}
\end{table*}
\end{CJK*}

The results in Table~\ref{tab:locale-degradation} highlight notable disparities in multilingual performance across LLMs. GPT-4o and Claude 3.7 Sonnet demonstrate stronger multilingual robustness, showing smaller average degradation from English (en\_US) to other locales. In contrast, GPT-4o-mini and Claude 3.5 Haiku suffer larger degradations—especially in Japanese (ja\_JP) and Spanish (es\_ES). DeepSeek-v3 shows promising overall stability, though with noticeable drops in East Asian languages. Across the board, Japanese (ja\_JP) and Chinese (zh\_CN) consistently show the largest degradation gaps, underscoring ongoing challenges in CJK locale generalization. GPT-o3 still leads in absolute en\_US performance, but newer models like GPT-4o and Claude 3.7 better balance quality and multilingual consistency.

\begin{table}[t]
\centering
\begin{adjustbox}{max width=\textwidth}
\begin{tabular}{l|c|cc|cc|cc|cc|cc}
\toprule
\multirow{2}{*}{\textbf{Model}} & \multirow{2}{*}{\textbf{\shortstack{en\_US\\(Base)}}} &
\multicolumn{2}{c|}{\textbf{es\_ES}} & \multicolumn{2}{c|}{\textbf{de\_DE}} &
\multicolumn{2}{c|}{\textbf{fr\_FR}} & \multicolumn{2}{c|}{\textbf{ja\_JP}} &
\multicolumn{2}{c}{\textbf{zh\_CN}} \\
\cline{3-12}
& & \textbf{Acc.} & \textbf{delta} & \textbf{Acc.} & \textbf{delta} & \textbf{Acc.} & \textbf{delta} & \textbf{Acc.} & \textbf{delta} & \textbf{Acc.} & \textbf{delta} \\
\midrule
GPT-o3            & \textbf{0.8309} & 0.7412 & \textcolor{red}{-8.97\%} & 0.7481 & \textcolor{red}{-8.28\%} & 0.7065 & \textcolor{red}{-12.44\%} & 0.6842 & \textcolor{red}{-14.67\%} & 0.7023 & \textcolor{red}{-12.86\%} \\
GPT-4o            & 0.7865 & 0.7055 & \textcolor{red}{-8.10\%} & 0.7235 & \textcolor{red}{-6.30\%} & 0.7032 & \textcolor{red}{-8.33\%} & 0.6326 & \textcolor{red}{-15.39\%} & 0.6690 & \textcolor{red}{-11.75\%} \\
GPT-4o-mini       & 0.7468 & 0.5913 & \textcolor{red}{-15.55\%} & 0.6493 & \textcolor{red}{-9.75\%} & 0.6678 & \textcolor{red}{-7.90\%} & 0.6028 & \textcolor{red}{-14.40\%} & 0.6208 & \textcolor{red}{-12.60\%} \\
Claude 3.7 Sonnet & 0.7876 & 0.6910 & \textcolor{red}{-9.66\%} & 0.7301 & \textcolor{red}{-5.75\%} & 0.6998 & \textcolor{red}{-8.78\%} & 0.6666 & \textcolor{red}{-12.10\%} & 0.7044 & \textcolor{red}{-8.32\%} \\
Claude 3.5 Haiku  & 0.7138 & 0.5697 & \textcolor{red}{-14.41\%} & 0.6840 & \textcolor{red}{-2.98\%} & 0.6271 & \textcolor{red}{-8.67\%} & 0.5658 & \textcolor{red}{-14.80\%} & 0.6220 & \textcolor{red}{-9.18\%} \\
DeepSeek-v3       & 0.7458 & 0.6617 & \textcolor{red}{-8.41\%} & 0.6797 & \textcolor{red}{-6.61\%} & 0.6923 & \textcolor{red}{-5.35\%} & 0.5812 & \textcolor{red}{-16.46\%} & 0.5997 & \textcolor{red}{-14.61\%} \\
\bottomrule
\end{tabular}
\end{adjustbox}
\caption{Macro accuracy and absolute degradation (delta) across locales, with \ en\_US as baseline.}
\label{tab:locale-degradation}
\end{table}

%% file: astra-bench.bib
@article{lu2024toolsandbox,
  title={Toolsandbox: A stateful, conversational, interactive evaluation benchmark for llm tool use capabilities},
  author={Lu, Jiarui and Holleis, Thomas and Zhang, Yizhe and Aumayer, Bernhard and Nan, Feng and Bai, Felix and Ma, Shuang and Ma, Shen and Li, Mengyu and Yin, Guoli and others},
  journal={arXiv preprint arXiv:2408.04682},
  year={2024}
}

@article{andrews2025scaling,
  title={Are: Scaling up agent environments and evaluations},
  author={Andrews, Pierre and Benhalloum, Amine and Bertran, Gerard Moreno-Torres and Bettini, Matteo and Budhiraja, Amar and Cabral, Ricardo Silveira and Do, Virginie and Froger, Romain and Garreau, Emilien and Gaya, Jean-Baptiste and others},
  journal={arXiv preprint arXiv:2509.17158},
  year={2025}
}

@article{barres2025tau,
  title = {$\tau^{2}$-Bench: Evaluating Conversational Agents in a Dual-Control Environment},
  author = {Barres, Victor and Dong, Honghua and Ray, Soham and Si, Xujie and Narasimhan, Karthik},
  journal = {arXiv preprint arXiv:2506.07982},
  year = {2025}
}

@article{wang2024gta,
  title={GTA: a benchmark for general tool agents},
  author={Wang, Jize and Zerun, Ma and Li, Yining and Zhang, Songyang and Chen, Cailian and Chen, Kai and Le, Xinyi},
  journal={Advances in Neural Information Processing Systems},
  volume={37},
  pages={75749--75790},
  year={2024}
}

@article{huang2024planning,
  title={Planning, creation, usage: Benchmarking llms for comprehensive tool utilization in real-world complex scenarios},
  author={Huang, Shijue and Zhong, Wanjun and Lu, Jianqiao and Zhu, Qi and Gao, Jiahui and Liu, Weiwen and Hou, Yutai and Zeng, Xingshan and Wang, Yasheng and Shang, Lifeng and others},
  journal={arXiv preprint arXiv:2401.17167},
  year={2024}
}

@article{wang2025rethinking,
  title={Rethinking Stateful Tool Use in Multi-Turn Dialogues: Benchmarks and Challenges},
  author={Wang, Hongru and Huang, Wenyu and Wang, Yufei and Xi, Yuanhao and Lu, Jianqiao and Zhang, Huan and Hu, Nan and Liu, Zeming and Pan, Jeff Z and Wong, Kam-Fai},
  journal={arXiv preprint arXiv:2505.13328},
  year={2025}
}

@article{backlund2025vending,
  title={Vending-bench: A benchmark for long-term coherence of autonomous agents},
  author={Backlund, Axel and Petersson, Lukas},
  journal={arXiv preprint arXiv:2502.15840},
  year={2025}
}

@inproceedings{mialon2023gaia,
  title={Gaia: a benchmark for general ai assistants},
  author={Mialon, Gr{\'e}goire and Fourrier, Cl{\'e}mentine and Wolf, Thomas and LeCun, Yann and Scialom, Thomas},
  booktitle={The Twelfth International Conference on Learning Representations},
  year={2023}
}

@article{mok2025exploring,
  title={Exploring the Potential of LLMs as Personalized Assistants: Dataset, Evaluation, and Analysis},
  author={Mok, Jisoo and Kim, Ik-hwan and Park, Sangkwon and Yoon, Sungroh},
  journal={arXiv preprint arXiv:2506.01262},
  year={2025}
}

@article{li2023api,
  title={Api-bank: A comprehensive benchmark for tool-augmented llms},
  author={Li, Minghao and Zhao, Yingxiu and Yu, Bowen and Song, Feifan and Li, Hangyu and Yu, Haiyang and Li, Zhoujun and Huang, Fei and Li, Yongbin},
  journal={arXiv preprint arXiv:2304.08244},
  year={2023}
}

@inproceedings{patil2025bfcl,
title={The Berkeley Function Calling Leaderboard (BFCL): From Tool Use to Agentic Evaluation of Large Language Models}, 
author={Patil, Shishir G. and Mao, Huanzhi and Cheng-Jie Ji, Charlie and Yan, Fanjia and Suresh, Vishnu and Stoica, Ion and E. Gonzalez, Joseph},
booktitle={Forty-second International Conference on Machine Learning},
year={2025},
}

@article{thoppilan2022lamda,
  title={Lamda: Language models for dialog applications},
  author={Thoppilan, Romal and De Freitas, Daniel and Hall, Jamie and Shazeer, Noam and Kulshreshtha, Apoorv and Cheng, Heng-Tze and Jin, Alicia and Bos, Taylor and Baker, Leslie and Du, Yu and others},
  journal={arXiv preprint arXiv:2201.08239},
  year={2022}
}

@article{liu2023agentbench,
  title={Agentbench: Evaluating llms as agents},
  author={Liu, Xiao and Yu, Hao and Zhang, Hanchen and Xu, Yifan and Lei, Xuanyu and Lai, Hanyu and Gu, Yu and Ding, Hangliang and Men, Kaiwen and Yang, Kejuan and others},
  journal={arXiv preprint arXiv:2308.03688},
  year={2023}
}

@article{farn2023tooltalk,
  title={Tooltalk: Evaluating tool-usage in a conversational setting},
  author={Farn, Nicholas and Shin, Richard},
  journal={arXiv preprint arXiv:2311.10775},
  year={2023}
}

@article{liu2025deepseek,
  title={Deepseek-v3.2: Pushing the frontier of open large language models},
  author={Liu, Aixin and Mei, Aoxue and Lin, Bangcai and Xue, Bing and Wang, Bingxuan and Xu, Bingzheng and Wu, Bochao and Zhang, Bowei and Lin, Chaofan and Dong, Chen and others},
  journal={arXiv preprint arXiv:2512.02556},
  year={2025}
}

@article{yao2024tau,
  title   = {{$\tau$}-Bench: A Benchmark for Tool–Agent–User Interaction in Real-World Domains},
  author  = {Yao, Shunyu and Shinn, Noah and Razavi, Pedram and Narasimhan, Karthik},
  journal = {arXiv preprint arXiv:2406.12045},
  year    = {2024}
}

@article{qin2023toolllm,
  title={Toolllm: Facilitating large language models to master 16000+ real-world apis},
  author={Qin, Yujia and Liang, Shihao and Ye, Yining and Zhu, Kunlun and Yan, Lan and Lu, Yaxi and Lin, Yankai and Cong, Xin and Tang, Xiangru and Qian, Bill and others},
  journal={arXiv preprint arXiv:2307.16789},
  year={2023}
}

@article{schick2023toolformer,
  title={Toolformer: Language Models Can Teach Themselves to Use Tools},
  author={Schick, Timo and Dwivedi-Yu, Arun Tejasvi and Hosseini, Sahana and Hou, Yuxuan and Schütze, Hinrich and Schlangen, David},
  journal={arXiv preprint arXiv:2302.04761},
  year={2023}
}

@article{chen2019bert,
  title={Bert for joint intent classification and slot filling},
  author={Chen, Qian and Zhuo, Zhu and Wang, Wen},
  journal={arXiv preprint arXiv:1902.10909},
  year={2019}
}

@article{brown2020language,
  title={Language models are few-shot learners},
  author={Brown, Tom and Mann, Benjamin and Ryder, Nick and Subbiah, Melanie and Kaplan, Jared D and Dhariwal, Prafulla and Neelakantan, Arvind and Shyam, Pranav and Sastry, Girish and Askell, Amanda and others},
  journal={Advances in neural information processing systems},
  volume={33},
  pages={1877--1901},
  year={2020}
}

@article{achiam2023gpt,
  title={Gpt-4 technical report},
  author={Achiam, Josh and Adler, Steven and Agarwal, Sandhini and Ahmad, Lama and Akkaya, Ilge and Aleman, Florencia Leoni and Almeida, Diogo and Altenschmidt, Janko and Altman, Sam and Anadkat, Shyamal and others},
  journal={arXiv preprint arXiv:2303.08774},
  year={2023}
}

@inproceedings{yao2022react,
  title={React: Synergizing reasoning and acting in language models},
  author={Yao, Shunyu and Zhao, Jeffrey and Yu, Dian and Du, Nan and Shafran, Izhak and Narasimhan, Karthik R and Cao, Yuan},
  booktitle={ICLR},
  year={2022}
}

@inproceedings{park2023generative,
  title={Generative agents: Interactive simulacra of human behavior},
  author={Park, Joon Sung and O'Brien, Joseph and Cai, Carrie Jun and Morris, Meredith Ringel and Liang, Percy and Bernstein, Michael S},
  booktitle={Proceedings of the 36th annual acm symposium on user interface software and technology},
  pages={1--22},
  year={2023}
}

@article{nakano2021webgpt,
  title={WebGPT: Browser-assisted question-answering with human feedback},
  author={Nakano, Reiichiro and Hilton, Jacob and Balaji, Suchir and others},
  journal={arXiv preprint arXiv:2112.09332},
  year={2021}
}

@inproceedings{chen2023places,
  title = "{PLACES}: Prompting Language Models for Social Conversation Synthesis",
  author = "Chen, Maximillian and Papangelis, Alexandros and Tao, Chenyang and Kim, Seokhwan and Rosenbaum, Andy and Liu, Yang and Yu, Zhou and Hakkani-Tur, Dilek",
  booktitle = "Findings of the Association for Computational Linguistics: EACL 2023",
  year = "2023",
  pages = "844--868",
  address = "Dubrovnik, Croatia",
  publisher = "Association for Computational Linguistics"
}

@inproceedings{zhang2018personalizing,
  title = "Personalizing Dialogue Agents: {I} have a dog, do you have pets too?",
  author = "Zhang, Saizheng and Dinan, Emily and Urbanek, Jack and Szlam, Arthur and Kiela, Douwe and Weston, Jason",
  booktitle = "Proceedings of the 56th Annual Meeting of the Association for Computational Linguistics (Volume 1: Long Papers)",
  year = "2018",
  pages = "2204--2213",
  address = "Melbourne, Australia",
  publisher = "Association for Computational Linguistics"
}

@misc{anthropic_claude,
  title        = {Claude Models and API},
  author       = {{Anthropic}},
  year         = {2024},
  howpublished = {\url{https://docs.anthropic.com/claude}},
  note         = {Accessed 2025}
}

@inproceedings{jang2023conversation,
  title = "Conversation Chronicles: Towards Diverse Temporal and Relational Dynamics in Multi-Session Conversations",
  author = "Jang, Jihyoung and Boo, Minseong and Kim, Hyounghun",
  booktitle = "Proceedings of the 2023 Conference on Empirical Methods in Natural Language Processing (EMNLP)",
  year = "2023",
  pages = "13584--13606",
  address = "Singapore",
  publisher = "Association for Computational Linguistics"
}

@article{silverman2006abm,
    author = {Silverman, Barry G. and Johns, Michael and Cornwell, Jason and O'Brien, Kevin},
    title = {Human Behavior Models for Agents in Simulators and Games: Part I: Enabling Science with PMFserv},
    journal = {Presence: Teleoperators and Virtual Environments},
    volume = {15},
    number = {2},
    pages = {139-162},
    year = {2006},
    month = {04},
    eprint = {https://direct.mit.edu/pvar/article-pdf/15/2/139/1624419/pres.2006.15.2.139.pdf},
}

@book{spencer2002playwright,
  title={The playwright's guidebook: An insightful primer on the art of dramatic writing},
  author={Spencer, Stuart},
  year={2002},
  publisher={Macmillan}
}

@article{hurst2024gpt,
  title={Gpt-4o system card},
  author={Hurst, Aaron and Lerer, Adam and Goucher, Adam P and Perelman, Adam and Ramesh, Aditya and Clark, Aidan and Ostrow, AJ and Welihinda, Akila and Hayes, Alan and Radford, Alec and others},
  journal={arXiv preprint arXiv:2410.21276},
  year={2024}
}

@article{bai2023qwen,
  title={Qwen technical report},
  author={Bai, Jinze and Bai, Shuai and Chu, Yunfei and Cui, Zeyu and Dang, Kai and Deng, Xiaodong and Fan, Yang and Ge, Wenbin and Han, Yu and Huang, Fei and others},
  journal={arXiv preprint arXiv:2309.16609},
  year={2023}
}

@article{trivedi2024appworld,
  title={Appworld: A controllable world of apps and people for benchmarking interactive coding agents},
  author={Trivedi, Harsh and Khot, Tushar and Hartmann, Mareike and Manku, Ruskin and Dong, Vinty and Li, Edward and Gupta, Shashank and Sabharwal, Ashish and Balasubramanian, Niranjan},
  journal={arXiv preprint arXiv:2407.18901},
  year={2024}
}
